\documentclass{article}

\usepackage{microtype}
\usepackage{graphicx}
\usepackage{subcaption}
\usepackage{booktabs}
\usepackage{multirow}
\usepackage{arydshln}
\usepackage{amsmath}
\usepackage{bm}

\usepackage{hyperref}

\usepackage[accepted]{icml2026}
\usepackage{amssymb}
\usepackage{mathtools}
\usepackage{amsthm}

\usepackage[capitalize,noabbrev]{cleveref}

\theoremstyle{plain}
\newtheorem{theorem}{Theorem}[section]
\newtheorem{proposition}[theorem]{Proposition}

\theoremstyle{definition}

\theoremstyle{remark}

\usepackage[textsize=tiny]{todonotes}

\icmltitlerunning{Spatiotemporal Imputation with Graph-Informed Flow Matching}

\begin{document}

\twocolumn[
  \icmltitle{Spatiotemporal Imputation with Graph-Informed Flow Matching}



  \icmlsetsymbol{equal}{*}

  \begin{icmlauthorlist}
    \icmlauthor{Zepeng Zhang}{epfl}
    \icmlauthor{Aref Einizade}{subparis}
    \icmlauthor{Jhony H. Giraldo}{telecom}
    \icmlauthor{Olga Fink}{epfl}
  \end{icmlauthorlist}

  \icmlaffiliation{epfl}{IMOS, EPFL, 1015 Lausanne, Switzerland}
  \icmlaffiliation{telecom}{LTCI, Télécom Paris, Institut Polytechnique de Paris, 91120 Palaiseau, France}
  \icmlaffiliation{subparis}{SAMOVAR, Télécom SudParis, Institut Polytechnique de Paris, 91120 Palaiseau, France}

  % \icmlcorrespondingauthor{Zepeng Zhang}{zepeng.zhang@epfl.ch}
  \icmlcorrespondingauthor{Olga Fink}{olga.fink@epfl.ch}

  \icmlkeywords{Machine Learning, ICML}

  \vskip 0.3in
]



\printAffiliationsAndNotice{}  % no special notice (required even if empty)

\begin{abstract}
Missing data is a common challenge in spatiotemporal systems, arising in applications such as air quality monitoring and urban traffic management. 
Traditional machine learning approaches, like recurrent and graph neural networks, rely on iterative propagation, which tends to accumulate errors over time and space. 
Recent diffusion-based methods mitigate error propagation but require iterative sampling and often depend on problem-agnostic Gaussian priors, limiting both efficiency and effectiveness.
To address these limitations, we propose \textbf{GiFlow}, a \emph{Graph-Informed Flow Matching} framework for spatiotemporal imputation. 
GiFlow replaces the typical Gaussian prior with a graph-informed prior constructed via spatiotemporal filtering of observable signals, which better aligns the source distribution to the target and thereby simplifies the generation trajectory.
The flow field is parameterized by a hybrid vector field model that integrates spatial attention, temporal attention, and spatiotemporal propagation, enabling joint modeling of spatial and temporal dependencies. 
% Unlike diffusion models, GiFlow is trained via direct regression and supports deterministic, few-step generation at inference. 
Extensive experiments on both synthetic and real-world datasets demonstrate that the proposed GiFlow outperforms the state-of-the-art approaches in spatiotemporal imputation. The code is available at https://github.com/zepengzhang/GiFlow.
\end{abstract}

\section{Introduction}
Spatiotemporal data characterizes both spatial and temporal information and is ubiquitous in domains such as environmental science, urban systems, and climate forecasting \citep{atluri2018spatio,wang2020deep}. 
In practice, spatiotemporal data is often incomplete due to sensor failures, transmission errors, or system instability \citep{yi2016st}. 
The incompleteness of spatiotemporal data compromises the reliability of subsequent analyses \citep{ma2024spatiotemporal,marisca2024graph}, motivating the need for robust spatiotemporal imputation techniques \citep{cao2018brits,cini2022filling}.

Early approaches to spatiotemporal imputation rely on statistical models that impose restrictive assumptions on the underlying data distribution, such as temporal smoothness levels, often failing to capture complex, nonlinear dependencies \citep{liu2023pristi,he2025filling}. 
Deep learning methods have been introduced to better exploit spatiotemporal correlations. 
Specifically, recurrent neural networks (RNNs) are used to capture temporal dependencies by propagating hidden states \citep{cao2018brits}, while graph neural networks (GNNs) are deployed to model spatial relationships over the underlying graph topology \citep{cini2022filling}. 
Despite their success, these models generally rely on iterative propagation across space and time, which can lead to error accumulation and information bottlenecks \citep{deng2024learning,he2025filling,cini2025graph}.
    
Generative models provide an alternative paradigm by inferring the entire data distribution in a non-autoregressive manner.
Unlike RNN/GNN-based models that propagate intermediate estimates step by step, generative models can perform imputation jointly and conditionally on all available observations, thereby avoiding the accumulation of errors during iterative propagation \citep{liu2019naomi, liu2023pristi,he2025filling}.
Among them, diffusion models have demonstrated remarkable success across various domains \citep{croitoru2023diffusion,yang2023diffusion,cao2024survey}, and recent works have adapted them for spatiotemporal imputation \citep{liu2023pristi,he2025filling}. 
However, diffusion models typically rely on the problem-agnostic Gaussian prior, presenting an absence of the available problem-specific structure.
Moreover, the sampling of diffusion models requires many iterative denoising steps, and the imputation often demands multiple sampling runs followed by averaging, limiting both efficiency and robustness when applied to large-scale spatiotemporal data.

Recent work has explored flow matching (FM) as a generalization of diffusion models, which follows a deterministic transport path \citep{lipman2023flow,albergo2023building,liu2023flow}. 
FM avoids stochastic noise injection, supports efficient deterministic sampling, and does not rely on Gaussian priors.
These characteristics make FM particularly attractive for conditional tasks such as imputation, where partial observations encode strong structural information.
The flexibility on prior selection allows FM to have shorter generative paths, which enhances generation performance \citep{tong2024improving}.
Building on these insights, we propose \textbf{GiFlow}, the first \emph{Graph-Informed Flow Matching} framework for spatiotemporal imputation.
Unlike existing diffusion-based methods that rely on problem-agnostic Gaussian priors \citep{liu2023pristi,he2025filling}, GiFlow constructs a graph-informed prior using spatiotemporal filtering of observable signals, simplifying generation trajectories.
Combined with a hybrid vector field integrating attention mechanisms and spatiotemporal propagation, our approach overcomes the limitations of iterative propagation in RNN- and GNN-based models, as well as the unstructured priors and inefficiency of diffusion-based methods.

Our contributions are summarized as follows:
\begin{itemize}
    \item We introduce GiFlow, a novel generative model for spatiotemporal imputation that integrates graph-informed priors into the flow matching framework. 
    \item We design a graph-informed prior based on adaptive spatiotemporal filtering. 
    Compared to the problem-agnostic Gaussian prior, this problem-tailored prior is more aligned with the target distribution and provably reduces transport cost.
    We also theoretically analyze the relationship between filtering factors and the receptive field in the spatiotemporal filtering process.
    \item We conduct extensive experiments on both synthetic and real-world datasets, demonstrating that the proposed GiFlow model achieves competitive or superior performance across diverse missing patterns and missing rates, outperforming state-of-the-art baselines.
\end{itemize}

\section{Preliminaries}

\subsection{Notations and Problem Definition}
\textbf{Notations.}
We use calligraphic letters like $\mathcal{X}$ to represent sets, uppercase bold letters like $\mathbf{X}$ to represent matrices, lowercase bold letters like $\mathbf{x}$ to represent vectors, and lowercase letters like $x$ to represent scalars.
We use $X_{ij}$ to represent the element in the $i$-th row and $j$-th column of $\mathbf{X}$.
$\mathbf{1}$ stands for the all-one matrix.
We denote by $|x|$, $\|\mathbf{x}\|$, and $\|\mathbf{X}\|$ the absolute value of $x$, the $\ell_2$-norm of $\mathbf{x}$, and the Frobenius norm of $\mathbf{X}$, respectively.
$\mathrm{vec}(\cdot)$ is the vectorization operation of a matrix.
$\mathrm{diag}(\mathbf{x})$ represents a matrix with its diagonal elements given by vector $\mathbf{x}$.
$\circ$ represents element-wise multiplication between matrices and $\oplus$ denotes the Kronecker sum operator between matrices.
% $\mathrm{div}_{\mathbf{X}}$ denotes the divergence with respect to all entries of $\mathbf{X}$.

\textbf{Graphs and Signals.}
%\paragraph{Spatiotemporal Imputation} 
Let spatiotemporal data be represented as a matrix $\mathbf{X}\in \mathbb{R}^{N \times R}$, where the $r$-th column of $\mathbf{X}$ denotes the signals observed at time $r$ across $N$ nodes (\textit{e.g.}, traffic sensors or air quality stations).
We denote by $\mathcal{R}=\{1,\ldots,R\}$ the set of timesteps.
The relationships among the nodes are captured by a graph $\mathcal{G} = (\mathcal{N}, \mathcal{E})$, with $\mathcal{N}$ being the set of nodes and $\mathcal{E}$ being the set of edges. 
Let $\mathbf{A} \in \mathbb{R}^{N \times N}$ denote the adjacency matrix, $\mathbf{D} = \mathrm{diag}(\mathbf{A1})$ the degree matrix, and $\mathbf{L} = \mathbf{D} - \mathbf{A}$ the Laplacian. 
For simplicity, we focus on one-dimensional signals, though the method generalizes to multi-dimensional signals.

\textbf{Spatiotemporal Imputation.}
We consider scenarios where some entries of $\mathbf{X}$ are missing. 
Define a binary mask $\mathbf{M} \in \{0,1\}^{N \times R}$ such that $M_{ir} = 1$ if the data on the node $i$ at time $r$ is observed, and $0$ otherwise. 
The incomplete observations are then given by $\mathbf{X} \circ \mathbf{M}$.
The task of spatiotemporal imputation is to estimate the missing entries based on the incomplete observations, leveraging both spatial dependencies across nodes and temporal dependencies across timesteps.

\subsection{Conditional Flow Matching}\label{sec:CFM}

FM learns a vector field that transports samples from a source distribution to a target distribution \citep{albergo2023building,lipman2023flow,liu2023flow}.
Let $\phi_t: [0,1] \times \mathbb{R}^d \to \mathbb{R}^d$ denote a step-dependent flow map  with $t$ being the flow step, that evolves $\mathbf{x}_0 \sim p_0$ to $\mathbf{x}_1 \sim p_1$ via the ordinary differential equation (ODE):
\begin{equation}
d\phi_t(\mathbf{x}) = u_t(\phi_t(\mathbf{x}))dt, \quad \phi_0(\mathbf{x}) = \mathbf{x}_0,
\end{equation}
where $u_t: [0,1] \times \mathbb{R}^d \to \mathbb{R}^d$ is a step-dependent vector field.
This induces a step-dependent probability density path $p_t$ through the push-forward operator $p_t = [\phi_t]_* p_0$ \citep{lipman2023flow}.

The FM objective seeks a trainable vector field $v_t(\cdot; \boldsymbol{\theta})$ that approximates $u_t$:
\begin{equation}
\mathcal{L}_{\mathsf{FM}}(\boldsymbol{\theta}) = \mathbb{E}_{t \sim \mathcal{U}[0,1], \mathbf{x} \sim p_t} \big[ | v_t(\mathbf{x}; \boldsymbol{\theta}) - u_t(\mathbf{x}) |^2 \big].
\end{equation}
This objective allows sampling from $p_1$ given a sample from $p_0$, as well as modeling continuous sample dynamics. However, it is generally intractable, as both $p_t$ and $u_t$ are unknown.

Conditional flow matching (CFM) \citep{lipman2023flow} provides a tractable alternative by approximating a conditional vector field $u_t(\mathbf{x} \mid \mathbf{z})$:
\begin{equation}
\begin{aligned}
\mathcal{L}_{\text{CFM}}(\boldsymbol{\theta})
= \mathbb{E}_{t \sim \mathcal{U}[0,1],\, \mathbf{z} \sim q(z),\, \mathbf{x} \sim p_t(\mathbf{x}|\mathbf{z})}
\big[ \\
\quad | v_t(\mathbf{x};\boldsymbol{\theta}) - u_t(\mathbf{x}|\mathbf{z}) |^2 \big].
\end{aligned}
\end{equation}

Here, $\mathbf{z}$ is chosen such that the marginal distributions of $p_t(\mathbf{x} \mid \mathbf{z})$ match the boundary distributions $p_0$ and $p_1$. 
Typically, $\mathbf{z} = (\mathbf{x}_0, \mathbf{x}_1)$ is sampled from a joint distribution $q(\mathbf{z}) = \pi(\mathbf{x}_0, \mathbf{x}_1)$ with marginals $p_0$ and $p_1$. Importantly, $\mathcal{L}_{\mathsf{CFM}}$ and $\mathcal{L}_{\mathsf{FM}}$ are equivalent in the sense that their gradients with respect to $\boldsymbol{\theta}$ coincide \citep{lipman2023flow}.

More discussions on spatiotemporal imputation and flow matching are provided in Appendix \ref{app:related-work}.
\begin{figure*}
    \centering
    \includegraphics[width=\textwidth]{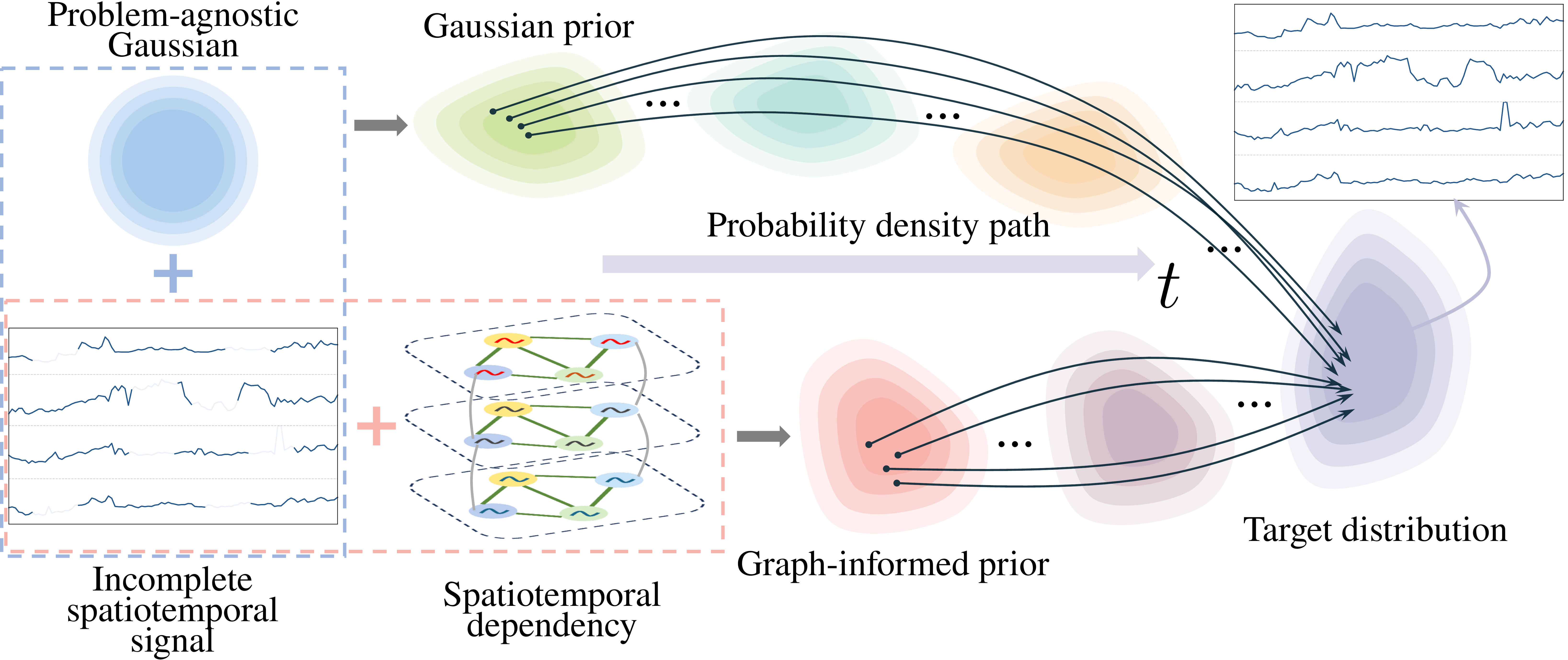}
    \caption{Schematic comparison of GiFlow and FM-Gauss (a FM model with a Gaussian prior).
    The red and blue dashed lines represent the information used to generate the prior for GiFlow and FM-Gauss, respectively.
    GiFlow constructs a graph-informed prior via adaptive spatiotemporal filtering of the observable signals, aligning the source distribution closer to the target, and hence simplifying the generation trajectories (lower part). 
    In contrast, because a problem-agnostic Gaussian prior ignores the spatiotemporal structure, it differs significantly from the target distribution, and thereby the model must traverse a longer path to reach the target distribution (upper part).}
    \label{fig:fm_framework}
\end{figure*}
\section{Graph-Informed Flow Matching}
In this section, we first construct a graph-informed prior via adaptive spatiotemporal filtering and then introduce the GiFlow framework for spatiotemporal imputation, with theoretical justifications on its effectiveness.
A schematic overview of the GiFlow framework is provided in Figure \ref{fig:fm_framework}.

\subsection{Graph-Informed Prior via Adaptive Spatiotemporal Filtering} 

Let $\mathbf{X}_1^M=\mathbf{X}_1\circ\mathbf{M}$ denote the observable spatiotemporal signal. 
The goal of GiFlow is to model the conditional distribution $p_1(\mathbf{X_1}\mid \mathbf{X}_1^M)$ via a generative model $p_{\boldsymbol{\theta}}(\mathbf{X_1}\mid \mathbf{X}_1^M)$.
Typically, generative models start from an isotropic Gaussian and treat $\mathbf{X}_1^M$ as a conditioning variable \citep{liu2023pristi,he2025filling}. 
However, the reliance on such a simple prior complicates the generative process as the prior distribution differs significantly from the target distribution \citep{kollovieh2025flow}.
To construct a more structured prior, we can decompose the conditional distribution as
\begin{equation*}
p_{\boldsymbol{\theta}}(\mathbf{X_1}\mid \mathbf{X}_1^M)
=\int p_{\boldsymbol{\theta}}(\mathbf{X_1}\mid \mathbf{X}_0, \mathbf{X}_1^M) q_0(\mathbf{X}_0 \mid \mathbf{X}_1^M) \, d\mathbf{X}_0.
\end{equation*}
Setting $q_0$ to a problem-agnostic standard Gaussian, as in most existing diffusion and FM models, neglects the spatiotemporal structure.
Leveraging the flexibility of FM, we construct a graph-informed prior more aligned to the target data distribution $p_1$.
By facilitating such alignment, we aim to reduce the overall transport cost.

We construct the graph-informed prior leveraging the joint continuous spatiotemporal filtering, which has been adopted in previous joint spatiotemporal frameworks \citep{stanley2020multiway,pan2021spatiotemporal,einizade2024continuous}.
Specifically, we consider $\mathbf{x}_1^M = \mathrm{vec}(\mathbf{X}_1^M)$ as a spatiotemporal graph signal living on top of the Cartesian product between the spatial and temporal graphs.
We denote by $\mathbf{L}_\eta$ and $\mathbf{L}_\xi$ the spatial and temporal graph Laplacians, respectively. 
Then the spatiotemporal filtering operator can be defined as the Kronecker sum of $\mathbf{L}_\eta$ and $\mathbf{L}_\xi$, \textit{i.e.}, $\mathbf{L}_{\eta \xi}=\tau_\xi \mathbf{L}_\xi\oplus\tau_\eta \mathbf{L}_\eta$, where $\tau_\eta$ and $\tau_\xi$ control the range of the receptive field.
In this way, the joint spatiotemporal filtering operation can be described as $\mathbf{x}_{\boldsymbol{\tau}}=e^{-\mathbf{L}_{\eta \xi}}\mathbf{x}_1^M$. 
It can be shown \citep{stanley2020multiway} that the matrix form of this spatiotemporal filtering operation, where $\mathbf{x}_{\boldsymbol{\tau}}=\mathrm{vec}(\mathbf{X}_{\boldsymbol{\tau}})$, takes the form of
\begin{equation}
\mathbf{X}_{\boldsymbol{\tau}} = e^{-\tau_\eta \mathbf{L}_\eta} \, \mathbf{X}_1^M \, e^{-\tau_\xi \mathbf{L}_\xi}.
\label{eq:graph_filtering}
\end{equation}
The continuous spatiotemporal model enables adaptive filtering. 
From the perspective of minimizing transport cost, we obtain the optimal $\boldsymbol{\tau}=(\tau_\eta,\tau_\xi)$ by solving the following minimization problem:
\begin{equation}
\begin{aligned}
&\underset{\tau_\eta,\tau_\xi>0}{\mathrm{minimize}}\;
\left\Vert \mathbf{X}_{1}-e^{-\tau_\eta\mathbf{L}_\eta}\mathbf{X}_1^Me^{-\tau_\xi\mathbf{L}_\xi}\right\Vert^2
+\\
&\alpha_\tau \mathrm{tr}\!\left( \big(e^{-\tau_\eta\mathbf{L}_\eta}\mathbf{X}_1^Me^{-\tau_\xi\mathbf{L}_\xi}\big)^\top \mathbf{L}_\eta \, e^{-\tau_\eta\mathbf{L}_\eta}\mathbf{X}_1^Me^{-\tau_\xi\mathbf{L}_\xi} \right),
\end{aligned}
\label{eq:tau-optimization}
\end{equation}
where the first term enforces signal alignment and the second term encourages Laplacian smoothness \citep{bontonou2019introducing,dong2020graph}.
This optimization balances alignment and smoothness, producing a spatiotemporal graph-informed prior that is close to the target distribution.

Expanding the exponentials via the Taylor series gives
\begin{equation}
\mathbf{X}_{\boldsymbol{\tau}}
= \left(\sum_{k=0}^{\infty}\frac{(-\tau_\eta)^k}{k!}\mathbf{L}_\eta^k\right) \mathbf{X}_1^M
  \left(\sum_{m=0}^{\infty}\frac{(-\tau_\xi)^m}{m!}\mathbf{L}_\xi^m\right),
\end{equation}
which propagates information across all nodes and timesteps for any nonzero $(\tau_\eta,\tau_\xi)$. 
Truncating it to $K_\eta$ spatial hops and $K_\xi$ temporal hops gives
\begin{equation*}
\mathbf{X}_{\boldsymbol{\tau}}^{K_\eta,K_\xi} =
\left(\sum_{k=0}^{K_\eta}\frac{(-\tau_\eta)^k}{k!}\mathbf{L}_\eta^k\right) \mathbf{X}_1^M
\left(\sum_{m=0}^{K_\xi}\frac{(-\tau_\xi)^m}{m!}\mathbf{L}_\xi^m\right).
\end{equation*}
\begin{proposition}[Adaptive spatiotemporal receptive field] \label{prop:adaptive_receptive_field}
Let $C_s$ and $C_t$ denote the spectral radii of $\mathbf{L}_\eta$ and $\mathbf{L}_\xi$, respectively. 
Then the truncation error is bounded by
\begin{equation}
\begin{aligned}
     &\left\|\mathbf{X}_{\boldsymbol{\tau}} - \mathbf{X}_{\boldsymbol{\tau}}^{K_\eta,K_\xi}\right\|\\\leq&\Bigg(\Bigg(\sum_{k=K_\eta+1}^{\infty}\frac{|\tau_\eta|^k}{k!} C_s^k \Bigg)\cdot
\left(\sum_{m=0}^{\infty}\frac{|\tau_\xi|^m}{m!} C_t^m \right)\\
     &+\left(\sum_{k=0}^{\infty}\frac{|\tau_\eta|^k}{k!} C_s^k \right)\cdot
\Bigg(\sum_{m=K_\xi+1}^{\infty}\frac{|\tau_\xi|^m}{m!} C_t^m \Bigg)\Bigg) \|\mathbf{X}_1^M\|.
\end{aligned}   
\label{eq:truncation_error}
\end{equation}
\end{proposition}
The proof of Proposition~\ref{prop:adaptive_receptive_field} is provided in Appendix~\ref{sec:proof_prop}.
According to Eq.~\eqref{eq:truncation_error}, the truncation error can be reduced either by decreasing $(\tau_\eta,\tau_\xi)$ or increasing $(K_\eta,K_\xi)$. In particular, for smaller filtering factor $(\tau_\eta,\tau_\xi)$, a smaller truncation order $(K_\eta, K_\xi)$ suffices to achieve the same approximation error. Therefore, $(\tau_\eta,\tau_\xi)$ effectively controls the spatial and temporal receptive fields: smaller values yield more localized receptive fields, while larger values expand them to capture long-range dependencies. Optimizing $\boldsymbol{\tau}=(\tau_\eta,\tau_\xi)$ thus enables an adaptive spatiotemporal receptive field.

In the following theorem, we explicitly show how the graph-informed prior in GiFlow enables more efficient transport compared to standard FM start from an isotropic Gaussian, highlighting the benefit of incorporating structural spatiotemporal knowledge.

\begin{theorem}[Control of transport cost]
\label{thm:transport-cost}
Consider flow matching for spatiotemporal imputation. 
Let $p_0^{\mathrm{G}}$ denote the graph-informed prior obtained via the spatiotemporal filtering operator defined in Eq. \eqref{eq:graph_filtering}, with $(\tau_\eta, \tau_\xi)$ being the optimal solution to Problem~\eqref{eq:tau-optimization} with $\alpha_\tau=0$, and let $p_0^{\mathrm{Gauss}}$ be the standard isotropic Gaussian prior. 
Denote by $q_1$ the target distribution.
Then, the transport cost of flow matching with $p_0^{\mathrm{G}}$ is no larger than that with $p_0^{\mathrm{Gauss}}$:
\begin{equation}
\mathcal{C}_{\mathrm{FM}}(p_0^{\mathrm{G}} \to q_1)
\leq
\mathcal{C}_{\mathrm{FM}}(p_0^{\mathrm{Gauss}} \to q_1),
\end{equation}
where $\mathcal{C}_{\mathrm{FM}}$ denotes the expected quadratic cost along the probability path.
\end{theorem}
The proof of Theorem~\ref{thm:transport-cost} is provided in Appendix~\ref{sec:proof_transport_cost}.
Intuitively, since the standard Gaussian prior ignores the spatiotemporal structure, the model must traverse a longer path to reach the target distribution. 
In contrast, the graph-informed prior integrates spatial smoothness and temporal consistency via adaptive spatiotemporal filtering, aligning the source distribution closer to the target distribution and thereby reducing the overall transport cost.

It is worth noting that since GiFlow adopts a deterministic input, it does not require multiple sampling runs followed by averaging as in existing diffusion approaches \citep{liu2023pristi,he2025filling}, which improves computational efficiency. 
Nonetheless, in scenarios where uncertainty quantification of the imputations is desired, additional Gaussian noise can be injected into the graph-informed prior to enable stochastic sampling.

\subsection{Graph-Informed Probability Flows}

For imputation problems, the data naturally comes in pairs $(\mathbf{X}_0, \mathbf{X}_1)$ \citep{albergo2024stochastic}.
To construct an FM model, it suffices to specify a conditional probability path and a vector field. 
We adopt a linear conditional probability path, which is optimal in the sense that the resulting conditional flow corresponds to the optimal transport displacement map, minimizing a bound on the kinetic energy \citep{lipman2023flow}.
Specifically, for a data pair $(\mathbf{X}_1^M, \mathbf{X}_1)$, the graph-informed linear conditional flow is defined as
\begin{equation}
\phi_t(\mathbf{X} \mid \mathbf{Z}) = (1-t) e^{-\tau_\eta \mathbf{L}_\eta} \mathbf{X}_1^M e^{-\tau_\xi \mathbf{L}_\xi} + t \mathbf{X}_1.
\label{eq:conditional_flow}
\end{equation}
This induces a unique vector field:
\begin{equation}
u_t(\mathbf{X} \mid \mathbf{Z}) = \mathbf{X}_1 - e^{-\tau_\eta \mathbf{L}_\eta} \mathbf{X}_1^M e^{-\tau_\xi \mathbf{L}_\xi}.
\end{equation}
Let $v_t$ be the parameterized vector field. The regression loss of GiFlow is then given by
\begin{equation*}
\begin{aligned}
&\mathcal{L}_\mathsf{GiFM}(\boldsymbol{\theta}) = \mathbb{E}_{t \sim \mathcal{U}[0,1], \mathbf{Z} \sim q(\mathbf{Z}), \mathbf{X} \sim p_t(\mathbf{X})} \Big[\\
& \left\Vert \mathbf{M} \circ \Big( v_t(\mathbf{X}_t; \boldsymbol{\theta}, \mathbf{M}, \mathbf{L}) - \mathbf{X}_1 + e^{-\tau_\eta \mathbf{L}_\eta} \mathbf{X}_1^M e^{-\tau_\xi \mathbf{L}_\xi} \Big) \right\Vert^2 \Big].
\end{aligned}
\end{equation*}

\subsection{Vector Field Model}
We parameterize the vector field model $v_t$ using a spatiotemporal model that captures both spatial and temporal dependencies, which has three main components: spatial attention, temporal attention, and spatiotemporal propagation.

\textbf{Spatial attention.} 
We first learn correlations between nodes using static node embeddings. 
Node embeddings are processed by a GNN developed in \citep{morris2019weisfeiler} to capture spatial information.
The propagated node embedding $\mathbf{X}_n$ serves as both the key and query for spatial attention.
The value is computed by $\mathbf{X}^{\eta}_t=\mathrm{MLP}(\mathbf{X}_t)\in\mathbb{R}^{N\times H}$, where $\mathbf{X}_t$ is computed using the defined conditional flow as in Eq. \eqref{eq:conditional_flow}.
To learn pairwise spatial associations, we employ self-attention:
\begin{align*}
&\mathbf{Q}^{\eta} = \mathbf{X}_n \mathbf{W}_Q^{\eta}, \quad
\mathbf{K}^{\eta} = \mathbf{X}_n \mathbf{W}_K^{\eta}, \quad
\mathbf{V}^{\eta} = \mathbf{X}^{\eta}_t \mathbf{W}_V^{\eta}, \\
&\alpha^{\eta}_{n_1,n_2} = \frac{\exp(\langle \mathbf{q}^{\eta}_{n_1}, \mathbf{k}^{\eta}_{n_2} \rangle)}{\sum_{n^\prime \in \mathcal{N}} \exp(\langle \mathbf{q}^{\eta}_{n_1}, \mathbf{k}^{\eta}_{n^\prime} \rangle)}, 
\end{align*}
where $\mathbf{W}_Q^{\eta}, \mathbf{W}_K^{\eta}, \mathbf{W}_V^{\eta} \in \mathbb{R}^{H \times H}$ are learnable matrices, $\mathbf{Q}^{\eta}, \mathbf{K}^{\eta}, \mathbf{V}^{\eta} \in \mathbb{R}^{N \times H}$ denote the query, key, and value matrices for spatial self-attention, with $\mathbf{q}^{\eta}_i, \mathbf{k}^{\eta}_i, \mathbf{v}^{\eta}_i$ representing their $i$-th rows. 
For a given spatiotemporal point $(n,r)$, we aggregate spatial messages from all nodes, weighted by the learned attention scores, to obtain the spatial embedding for each node. 
This aggregation is computed as
\begin{equation}
\mathbf{h}^{\eta}_{n} = \mathrm{MLP}\Bigg( \sum_{n^\prime \in \mathcal{N}} \alpha^{\eta}_{n,n^\prime} , \mathbf{v}^{\eta}_{n^\prime} \Bigg).
\end{equation}

\textbf{Temporal attention.} 
To capture correlations across timesteps, we employ a temporal attention mechanism. 
Unlike recurrent sequence models such as RNNs or LSTMs, Transformers do not inherently encode sequential information \citep{wen2023transformers}. 
To address this, we first incorporate standard positional encoding \citep{vaswani2017attention}. 
When real-world timestamps are available, we additionally use a learnable embedding layer to encode them \citep{zhou2021informer}.
Let $\mathbf{X}_{PE}$ and $\mathbf{X}_{TE}$ denote the positional encoding and timestamp encoding. 
The input to the temporal attention module is then given by
\begin{equation}
\mathbf{X}_{t}^{\xi} = \mathrm{MLP}(\mathbf{X}_{t}^\top) + \mathbf{X}_{PE} + \mathbf{X}_{TE} \in \mathbb{R}^{R \times H}.
\end{equation}

For any pair of timesteps $(r_1, r_2) \in \mathcal{R}$, $\mathbf{X}_t^{\xi}$ serves as the key, query, and value in the temporal attention computation. The temporal attention scores and aggregation are computed as follows:
\begin{align*}
& \mathbf{Q}^{\xi} = \mathbf{X}_t^{\xi} \mathbf{W}_Q^{\xi}, \quad
\mathbf{K}^{\xi} = \mathbf{X}_t^{\xi} \mathbf{W}_K^{\xi}, \quad
\mathbf{V}^{\xi} = \mathbf{X}_t^{\xi} \mathbf{W}_V^{\xi}, \\
& \alpha^{\xi}_{r_1,r_2} = \frac{\exp\big(\langle \mathbf{q}^{\xi}_{r_1}, \mathbf{k}^{\xi}_{r_2} \rangle\big)}
{\sum_{r^\prime \in \mathcal{R}} \exp\big(\langle \mathbf{q}^{\xi}_{r_1}, \mathbf{k}^{\xi}_{r^\prime} \rangle\big)},
\end{align*}
where $\mathbf{W}_Q^{\xi}, \mathbf{W}_K^{\xi}, \mathbf{W}_V^{\xi} \in \mathbb{R}^{H \times H}$ are learnable parameter matrices, and $\mathbf{Q}^{\xi}, \mathbf{K}^{\xi}, \mathbf{V}^{\xi} \in \mathbb{R}^{R \times H}$ denote the query, key, and value matrices. 
For a spatiotemporal point $(n, r)$, temporal messages from all timesteps are aggregated using the learned attention weights, producing the temporal embedding computed as
\begin{equation}
\mathbf{h}^{\xi}_{r} = \mathrm{MLP}\Bigg( \sum_{r^\prime \in \mathcal{R}} \alpha^{\xi}_{r, r^\prime} , \mathbf{v}^{\xi}_{r^\prime} \Bigg).
\end{equation}

\textbf{Spatiotemporal propagation.} 
The aggregated spatial and temporal messages are concatenated with the original features and time embedding that encodes the information of the step $t$ in the probability density path, then projected via a linear layer to obtain $\mathbf{H} \in \mathbb{R}^{N \times R \times H}$.
We then perform $L_{MP}$ layers of message passing in both spatial and temporal domains, with the $\ell$-th ($\ell=1,\ldots,L_{MP}$) layer defined by
\begin{equation}
\begin{aligned}
\mathbf{H}_r^{(\ell+1)} &= \mathrm{GNN}(\mathbf{H}_r, \mathcal{G}_s) \in \mathbb{R}^{N \times H}, \quad \forall r \in \mathcal{R}, \\
\mathbf{H}_n^{(\ell+1)} &= \mathrm{GNN}(\mathbf{H}_n, \mathcal{G}_t) \in \mathbb{R}^{R \times H}, \quad \forall n \in \mathcal{N},
\end{aligned}
\end{equation}
where spatial message passing is applied independently for each timestep, and temporal message passing is applied independently for each node. 
The GNN model follows the architecture developed in \citep{wu2019simplifying}.
After $L_{MP}$ layers, we obtain $\mathbf{H}^{prop} \in \mathbb{R}^{N \times R \times H}$.
Then a linear layer projects the features back to the original signal dimension.  
For one-dimensional signals, the final output is
\begin{equation}
\mathbf{X}^{out} = \mathrm{MLP}(\mathbf{H}^{prop}) \in \mathbb{R}^{N \times R}.
\end{equation}

The GiFlow model integrates graph-informed priors with a spatiotemporal architecture in the flow matching framework, providing an effective generative model for spatiotemporal imputation. 

\section{Experiments}
We assess the performance of GiFlow using synthetic data \citep{qiu2017time,giraldo2022reconstruction} as well as three widely used real-world datasets that have different sizes and spatiotemporal patterns: two air quality datasets (Air-36 and AQI) \citep{zheng2015forecasting, yi2016st} and a traffic dataset (PeMS08) \citep{guo2021learning}.
Specifically, Air-36 and AQI collect hourly sampled PM2.5 pollutant data in China.
PeMS08 is collected by the Caltrans Performance Measurement System (PeMS) \citep{chen2001freeway}, containing highway traffic flow data in California.
It originally collects data every 30 seconds, and the collected data is then aggregated with a 5-minute interval.
To simulate realistic incomplete spatiotemporal signals, we adopt two missing data injection strategies:
(1) Point missing: following the setup of \citep{cini2022filling,deng2024learning}, we randomly mask a fraction $\rho$ of the available data;
(2) Block missing: we first randomly select a node and a starting timestep, then mask a contiguous segment of data from that timestep for the selected node. This process is repeated iteratively until a fraction $\rho$ of the available data is masked.
We compare the performance of GiFlow with five non-parametric methods (Mean-S, Mean-T, Linear, KNN, and FP \citep{rossi2022on}), two RNN-based methods (BRITS \citep{cao2018brits} and SAITS \citep{du2023saits}), four spatiotemporal GNN-based and transformer-based methods (SPIN \citep{marisca2022learning}, GRIN \citep{cini2022filling}, OPCR \citep{deng2024learning}, and a diffusion-based method PriSTI \citep{liu2023pristi}.
To obtain the filtering factors $\tau_\eta$ and $\tau_\xi$, we optimize Problem \eqref{eq:tau-optimization} with stochastic gradient descent.
Specifically, the filtering factors are optimized using the training data, where the complete ground-truth signals are available. Once selected, the filtering factors remain constant during inference.
Details about the datasets and baselines are provided in Appendix \ref{app:datasets} and Appendix \ref{app:baselines}, respectively.
To evaluate the performance, we use three metrics: mean absolute error (MAE), root mean squared error (RMSE), and mean absolute percentage error (MAPE).
All the experiment results are conducted five times using different seeds, and we report the average performance.
The implementation details can be found in Appendix \ref{app:implementation_details}.
\begin{table}[t]
\centering
\caption{Imputation performance on the synthetic dataset.}
\label{table:synthetic}
\resizebox{\columnwidth}{!}{
\begin{tabular}{lcccccc}
\toprule
\multirow{2}{*}{Model} & \multicolumn{3}{c}{$\sigma=0.1$} & \multicolumn{3}{c}{$\sigma=0.3$} \\
\cmidrule(l){2-4} \cmidrule(l){5-7}
 & MAE & RMSE & MAPE & MAE & RMSE & MAPE\\
\midrule
BRITS  & 0.35 & 0.56 & 7.90  & 0.39 & 0.64 & 12.26 \\
SAITS  & 0.30 & 0.41 & 7.52  & 0.36 & 0.47 & 11.27 \\
SPIN   & 0.83 & 1.08 & 26.83 & 0.87 & 1.12 & 29.82 \\
GRIN   & \underline{0.24} & \underline{0.31} & \textbf{5.98}  
       & \underline{0.35} & \underline{0.46} & \underline{11.05} \\
OPCR   & 0.32 & 0.42 & 11.12 & 0.44 & 0.57 & 14.88 \\
PriSTI & 0.32 & 0.36 & 11.21 & 0.37 & 0.47 & 12.48 \\
[0.2em]
\cdashline{1-7}\\[-0.8em]
GiFlow & \textbf{0.23} & \textbf{0.30} & \underline{6.65}  
       & \textbf{0.34} & \textbf{0.44} & \textbf{10.67} \\
\bottomrule
\end{tabular}
}
\end{table}

\subsection{Performance Evaluation on Synthetic Datasets}\label{sec:synthetic}
\begin{table*}[t]
\centering
\caption{Imputation performance with point missing strategy ($\rho=20\%$).}
\label{tab:air-point}
{
\resizebox{0.82\textwidth}{!}{
\begin{tabular}{lcccccc}
\toprule
\multirow{2}{*}{Model} & \multicolumn{3}{c}{Air-36} & \multicolumn{3}{c}{AQI} \\
\cmidrule(l){2-4} \cmidrule(l){5-7} 
 & MAE & RMSE & MAPE & MAE & RMSE & MAPE \\
\midrule
Mean-S & $19.22_{\pm 0.17}$ & $31.81_{\pm 0.50}$ & $45.60_{\pm 0.75}$
    & $34.93_{\pm 0.05}$ & $48.94_{\pm 0.53}$ & $112.45_{\pm 0.27}$\\
Mean-T & $30.39_{\pm 0.22}$ & $44.83_{\pm 0.55}$ & $83.64_{\pm 1.22}$
    & $20.57_{\pm 0.02}$ & $33.78_{\pm 0.31}$ & $55.44_{\pm 0.18}$\\
Linear & $11.02_{\pm 0.12}$ & $21.28_{\pm 0.60}$ & $27.68_{\pm 0.50}$
    & $8.97_{\pm 0.04}$ & $19.95_{\pm 0.25}$ & $21.42_{\pm 0.11}$\\
KNN & $20.33_{\pm 0.14}$ & $34.25_{\pm 0.71}$ & $41.74_{\pm 0.96}$
    & $18.95_{\pm 0.03}$ & $33.03_{\pm 0.27}$ & $52.44_{\pm 0.20}$\\ 
FP & $16.51_{\pm 0.17}$ & $28.68_{\pm 0.80}$ & $38.51_{\pm 1.70}$
    & $15.65_{\pm 0.02}$ & $27.20_{\pm 0.41}$ & $45.70_{\pm 0.18}$\\
BRITS & $14.23_{\pm 0.17}$ & $24.64_{\pm 0.59}$ & $31.42_{\pm 0.72}$
    & $16.55_{\pm 0.12}$ & $26.78_{\pm 0.36}$ & $41.25_{\pm 0.35}$\\
SAITS & $14.32_{\pm 0.13}$ & $23.85_{\pm 0.61}$ & $31.62_{\pm 0.85}$
    & $17.95_{\pm 0.07}$ & $28.95_{\pm 0.41}$ & $44.89_{\pm 0.30}$\\
SPIN & $11.05_{\pm 0.87}$ & $20.97_{\pm 1.90}$ & $22.26_{\pm 1.56}$
    & $8.72_{\pm 0.12}$ & $19.61_{\pm 0.67}$ & $18.55_{\pm 0.28}$\\
GRIN & $\underline{9.94_{\pm 0.12}}$ & $\underline{19.09_{\pm 0.87}}$ & $21.95_{\pm 0.22}$ 
    & $\underline{7.97_{\pm 0.08}}$ & $\underline{18.46_{\pm 0.54}}$ & $16.81_{\pm 0.24}$\\
OPCR & $10.03_{\pm 0.12}$ & $19.32_{\pm 0.60}$ & $\underline{21.61_{\pm 0.68}}$ 
    & $8.40_{\pm 0.20}$ & $19.30_{\pm 0.69}$ & $16.91_{\pm 0.59}$\\
PriSTI & $10.29_{\pm 0.14}$ & $19.66_{\pm 0.25}$ & $21.91_{\pm 0.66}$ 
    & $8.17_{\pm 0.28}$ & $19.85_{\pm 0.28}$ & $\underline{16.37_{\pm 0.59}}$\\
CoSTI & $9.95_{\pm 0.10}$ & $19.66_{\pm 0.27}$ & $22.08_{\pm 0.37}$ 
    & $8.10_{\pm 0.13}$ & $18.59_{\pm 0.37}$ & $17.17_{\pm 0.41}$\\
[0.2em]
\cdashline{1-7}\\[-0.8em]
GiFlow & \boldsymbol{$9.54_{\pm 0.18}$} & \boldsymbol{$18.10_{\pm 0.78}$} & \boldsymbol{$21.27_{\pm 0.33}$} 
    & \boldsymbol{$7.83_{\pm 0.10}$} & \boldsymbol{$17.80_{\pm 0.28}$} & \boldsymbol{$16.24_{\pm 0.31}$}\\
\bottomrule
\end{tabular}
}}
\end{table*}
\begin{table*}[t]
\centering
\caption{Imputation performance with block missing strategy ($\rho=20\%$).}
\label{tab:air-block}
{
\resizebox{0.82\textwidth}{!}{
\begin{tabular}{lcccccc}
\toprule
\multirow{2}{*}{Model} & \multicolumn{3}{c}{Air-36} & \multicolumn{3}{c}{AQI} \\
\cmidrule(l){2-4} \cmidrule(l){5-7} 
 & MAE & RMSE & MAPE & MAE & RMSE & MAPE \\
\midrule

Mean-S & $19.80_{\pm 0.29}$ & $32.03_{\pm 0.23}$ & $44.00_{\pm 1.88}$
       & $34.96_{\pm 0.19}$ & $48.70_{\pm 0.91}$ & $117.44_{\pm 1.95}$ \\

Mean-T & $41.06_{\pm 0.42}$ & $59.70_{\pm 1.00}$ & $102.80_{\pm 2.19}$
       & $26.81_{\pm 0.13}$ & $42.69_{\pm 0.38}$ & $76.71_{\pm 1.19}$ \\

Linear & $33.03_{\pm 0.57}$ & $52.53_{\pm 1.28}$ & $81.95_{\pm 5.11}$
       & $23.87_{\pm 0.24}$ & $40.83_{\pm 0.75}$ & $69.03_{\pm 1.12}$ \\

KNN    & $20.78_{\pm 0.39}$ & $34.33_{\pm 0.56}$ & $40.34_{\pm 2.57}$
       & $18.58_{\pm 0.12}$ & $32.80_{\pm 0.36}$ & $52.55_{\pm 1.13}$ \\

FP     & $16.65_{\pm 0.30}$ & $28.66_{\pm 0.74}$ & $35.76_{\pm 2.02}$
       & $15.19_{\pm 0.14}$ & $26.74_{\pm 0.53}$ & $44.81_{\pm 1.08}$ \\

BRITS  & $17.78_{\pm 0.46}$ & $28.16_{\pm 0.93}$ & $32.19_{\pm 2.23}$
       & $17.60_{\pm 0.09}$ & $28.24_{\pm 0.19}$ & $44.21_{\pm 0.21}$ \\

SAITS  & $18.04_{\pm 0.23}$ & $28.63_{\pm 0.52}$ & $40.54_{\pm 1.01}$
       & $19.45_{\pm 0.16}$ & $30.85_{\pm 0.66}$ & $49.86_{\pm 0.79}$ \\

SPIN   & $16.59_{\pm 0.31}$ & $28.07_{\pm 0.37}$ & $31.17_{\pm 1.04}$
       & $14.73_{\pm 0.15}$ & $26.79_{\pm 0.47}$ & $32.23_{\pm 0.21}$ \\

GRIN   & $16.27_{\pm 0.32}$ & $27.67_{\pm 0.67}$ & $31.86_{\pm 1.17}$
       & $\underline{14.47_{\pm 0.14}}$ & $\mathbf{25.23_{\pm 0.23}}$ & $33.03_{\pm 0.32}$ \\

OPCR   & $15.27_{\pm 0.19}$ & $\underline{25.44_{\pm 1.10}}$ & $33.38_{\pm 1.03}$
       & $14.52_{\pm 0.28}$ & $25.95_{\pm 1.52}$ & $\underline{31.75_{\pm 0.54}}$ \\

PriSTI & $\underline{15.07_{\pm 0.65}}$ & $25.57_{\pm 1.27}$ & $\underline{29.84_{\pm 1.96}}$
       & $14.54_{\pm 0.45}$ & $26.61_{\pm 1.48}$ & $31.77_{\pm 0.39}$ \\
CoSTI & $15.19 _{\pm 0.27}$ & $26.85 _{\pm 0.79}$ & $30.85 _{\pm 0.63}$
    & $14.78 _{\pm 0.37}$ & $26.71 _{\pm 0.74}$ & $32.36 _{\pm 0.62}$ \\
[0.2em]
\cdashline{1-7}\\[-0.8em]
GiFlow & $\mathbf{14.76_{\pm 0.38}}$ & $\mathbf{25.33_{\pm 2.14}}$ & $\mathbf{28.95_{\pm 0.80}}$
       & $\mathbf{13.74_{\pm 0.30}}$ & \underline{$25.43_{\pm 2.61}$} & $\mathbf{31.09_{\pm 1.79}}$ \\
\bottomrule
\end{tabular}
}}
\end{table*}
To evaluate the proposed GiFlow, we generate a synthetic spatiotemporal dataset following the procedure described in \citep{qiu2017time,giraldo2022reconstruction}. 
This yields a smooth, temporally evolving graph signal $\mathbf{X} \in \mathbb{R}^{N \times R}$. 
Specifically, we sample 50 nodes uniformly at random within a 50$\times$50 square domain.
A graph is then constructed using KNN based on the spatial positions, with $k=5$.
Let $\mathbf{L}_\eta \in \mathbb{R}^{N \times N}$ be the spatial graph Laplacian, for which we compute the eigen-decomposition $\mathbf{L}_\eta = \mathbf{V} \mathbf{\Lambda} \mathbf{V}^\top$.
Its inverse square root is used to construct a smoothed propagation operator $\mathbf{L}_\eta^{-\frac{1}{2}}=\mathbf{U}\boldsymbol{\Lambda}^{-\frac{1}{2}}\mathbf{U}^\top$ where $\boldsymbol{\Lambda}^{-\frac{1}{2}}=\mathrm{diag}(0,\lambda_2^{-\frac{1}{2}},\ldots,\lambda_N^{-\frac{1}{2}})$.
The initial signal is generated in the spectral domain as a low-frequency signal. 
Subsequent signals are generated iteratively via $\mathbf{x}_r=\mathbf{x}_{r-1}+\mathbf{L}_\eta^{-\frac{1}{2}}\mathbf{f}_r$, where $\mathbf{f}$ is an i.i.d. Gaussian signal.
The length of the generated signal is set to $R=3000$.
To simulate the real-world noisy conditions, we add small Gaussian noise to the signal using
$
\tilde{\mathbf{X}} = \mathbf{X} + \boldsymbol{\epsilon}, \,
\boldsymbol{\epsilon} \sim \mathcal{N}(\mathbf{0}, \sigma^2 \mathbf{I}).
$
We conduct experiments using the point missing strategy ($\rho=20\%$) with both $\sigma=0.1$ and $\sigma=0.3$ to evaluate the performance under different noisy levels.
The results are given in Table \ref{table:synthetic}, where the best results are in \textbf{bold}, and the second best results are \underline{underlined}.
From the results, we can see that the GiFlow model performs well under different noisy levels.

\begin{figure}
    \centering
    \includegraphics[width=\columnwidth]{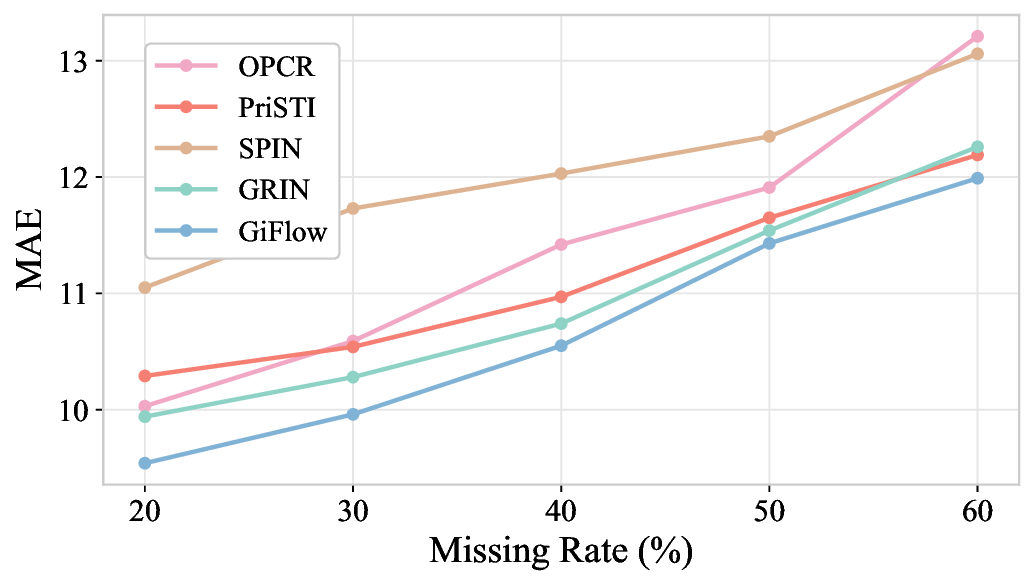}
    \includegraphics[width=\columnwidth]{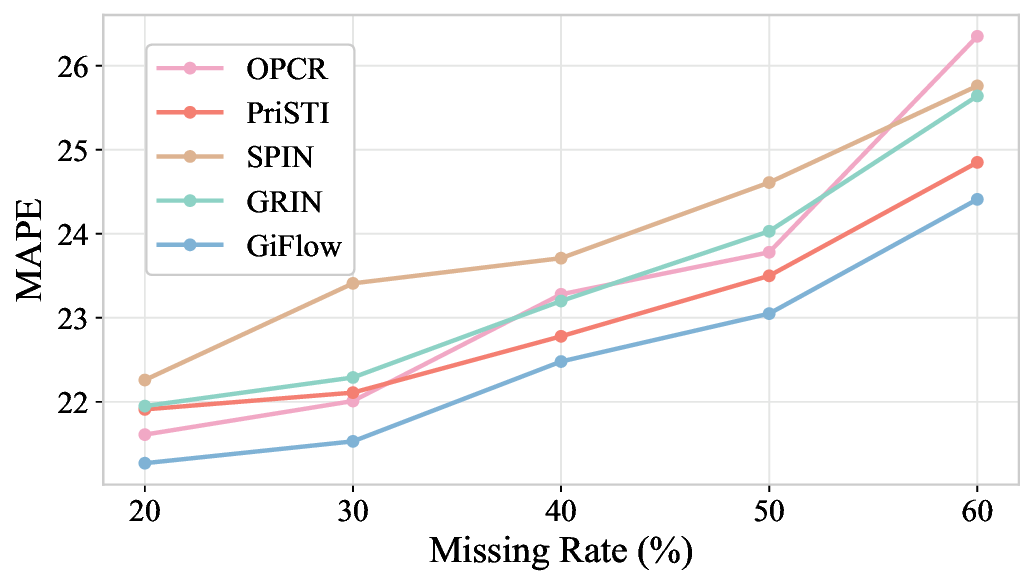}
    \caption{Performance with different missing rates.}
    \label{fig:missing-rate}
\end{figure}

\subsection{Performance Evaluation on Real-world Datasets}
We first evaluate model performance on the two air quality datasets: a small Air-36 dataset that is collected in Beijing and a large dataset that is collected from 43 Chinese cities.
The temporal graph is defined as a line graph to model the auto-regressive nature of the time series.
The spatial graph is constructed from training data using distance-based similarity.
Specifically, we first compute the pairwise distance matrix $\boldsymbol{\Psi}\in\mathbb{R}^{N\times N}$ using node features.
Then we apply a Gaussian kernel $\exp(-(\frac{\psi_{ij}}{\omega})^2)$ with the decay rate $\omega$ set to the standard deviation of $\boldsymbol{\Psi}$ to transform the distances into edge weights between 0 and 1.
The weight matrix is further converted into a binary adjacency matrix with thresholding.
The results with the point missing and block missing strategies ($\rho=20\%$) are reported in Table \ref{tab:air-point} and \ref{tab:air-block}, respectively.
% Most baselines show significant differences on both datasets, confirming that our gains are meaningful.
% For example, all methods show significantly lower performance on Air-36 in Table \ref{tab:air-point}.
The results show that simple averaging methods, Mean-S and Mean-T, fail to achieve satisfactory results, indicating that simple spatial or temporal averaging cannot capture the system dynamics. 
Linear interpolation performs well under point missing, but degrades significantly under block missing.
The same happens to RNN-based methods BRITS and SAITS, which even underperform the nonparametric FP model.
This is because for block missing, there are contiguous missing blocks, making methods that only rely on individual time series struggle with inferring missing values based on signals from distant timesteps.
The other deep learning methods that consider both spatial and temporal information, \textit{i.e.}, SPIN, GRIN, OPCR, PriSTI, and GiFlow, perform better than the methods using only temporal information.
Among them, GiFlow achieves the overall best results across different missing patterns and metrics, demonstrating its effectiveness in spatiotemporal imputation.

To further evaluate the effectiveness of the proposed GiFlow model, we conduct experiments on the Air-36 dataset, using the point missing strategy with $\rho$ ranging from 20\% to 60\%.
For comparison, we choose four spatiotemporal baselines, which are shown to be better than other baselines according to the results reported in Table \ref{tab:air-point} and \ref{tab:air-block}.
The results on MAPE are presented in Figure \ref{fig:missing-rate}, while the results on MAE and RMSE are presented in Appendix \ref{app:missing_rate}.
From the results, we observe that for all the methods, the imputation performance steadily degrades with increasing missing rates.
Moreover, the proposed GiFlow model consistently outperforms the other methods across different missing rates. 
These results validate the robustness of the proposed GiFlow model under different missing patterns and missing rates.

\begin{figure}
    \centering
    \includegraphics[width=\columnwidth]{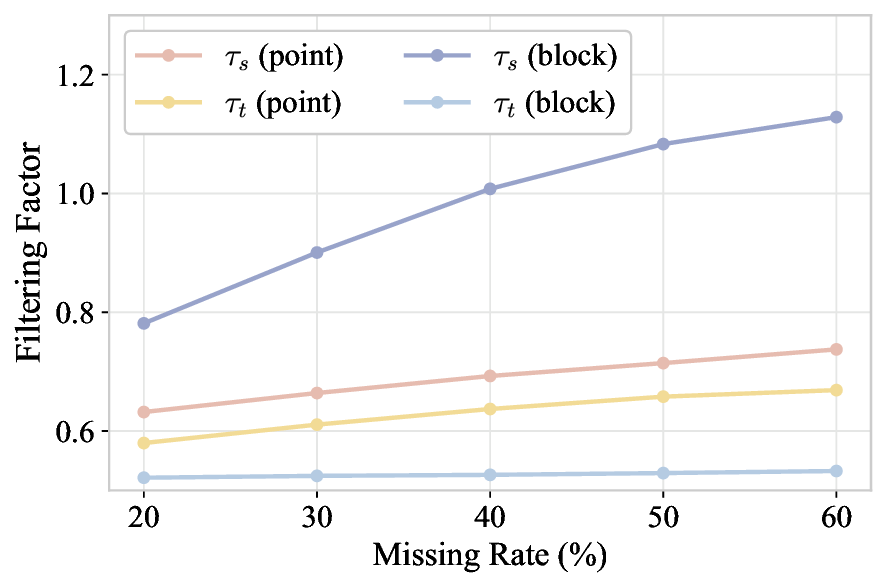}
    \caption{Filtering factors with different missing rate.}
    \label{fig:tau-missing}
\end{figure}
As proved in Proposition \ref{prop:adaptive_receptive_field}, the filtering factors $\tau_\eta$ and $\tau_\xi$ determine the receptive field of the spatiotemporal filtering operation. 
Intuitively, when facing a higher missing rate, the model would require a larger receptive field to obtain a close approximation of the missing signals based on the observable ones.
In the following, we investigate how $\tau_\eta$ and $\tau_\xi$ change as the missing rate $\rho$ increases.
The values of the filtering factors $\tau_\eta$ and $\tau_\xi$ under both the point missing strategy and block missing strategy on the Air-36 dataset with $\rho$ ranging from 20\% to 60\% are presented in Figure \ref{fig:tau-missing}.
It is evident that with increasing missing rate, $\tau_\eta$ and $\tau_\xi$ become larger and larger. 
This pattern corroborates the results we obtained from Proposition \ref{prop:adaptive_receptive_field} that with increasing missing rate, the model likely requires a larger spatiotemporal receptive field.
Moreover, for block missing, $\tau_\eta$ increases a lot as the missing rate increases, while $\tau_\xi$ remains relatively stable.
This is largely due to the fact that with the block missing strategy, we have larger temporal gaps in the data, making the model more reliant on spatial filtering.
A similar phenomenon has been observed in the AQI dataset, which is presented in Appendix \ref{app:tau_value}.

\subsection{Ablation Study}
\textbf{Ablation on prior construction.}
In the proposed GiFlow model, the graph-informed prior construction is critical as it provides a close alignment between the source and target distribution, and hence reduces the transport cost.
To empirically validate the effectiveness of the graph-informed prior generated with adaptive spatiotemporal filtering, we evaluate several variants of GiFlow.
Specifically, we consider: (1) FM-Gauss: a FM model employs the same vector field model architecture as in GiFlow but with a problem-agnostic Gaussian prior; (2) GFM: GiFlow with a spatial-only graph-informed prior, \textit{i.e.}, the filtering parameter $\tau_\eta$ is obtained by optimizing Problem \eqref{eq:tau-optimization} with $\tau_\xi$ fixed as 0; (3) TFM: GiFlow with a temporal-only graph-informed prior, \textit{i.e.}, the filtering parameter $\tau_\xi$ is obtained by optimizing Problem \eqref{eq:tau-optimization} with $\tau_\eta$ fixed as 0.
We conduct experiments on Air-36 dataset with point missing strategy.
To validate the results in Theorem \ref{thm:transport-cost}, we also evaluate the transport cost of different models.
The results are reported in Table \ref{tab:ablation}.
From the results, we observe that the FM-Gauss performs the worst, and it is worse than several baselines in Table \ref{tab:air-point}, emphasizing that FM with a Gaussian prior fails to achieve state-of-the-art imputation performance.
TFM and GFM give better results than FM-Gauss, indicating that the Laplacian filtering in both the spatial and temporal domains provides more structured and informative prior to the FM framework.
Notably, GFM outperforms all baselines in Table \ref{tab:air-point}, indicating that the spatial filtering alone already provides a good prior for the model. 
Leveraging on both spatial and temporal dependencies, GiFlow gives the best results, indicating that combining both spatial and temporal filtering brings additional performance gain.
The results on transport cost corroborate the conclusions of Theorem \ref{thm:transport-cost}. They demonstrate that applying graph filtering can substantially reduce the transport cost. Moreover, it can also be observed that models with lower transport costs tend to achieve better performance.

\begin{table}[t]
\centering
\caption{The effect of different priors. (T. C. stands for transport cost.)}
\label{tab:ablation}
{
\resizebox{\columnwidth}{!}{
\begin{tabular}{lcccc}
\toprule
Model & T. C. & MAE & RMSE & MAPE\\
\midrule
FM-Gauss 
    & 299.62
    & 12.79$_{\pm 0.63}$ 
    & 22.15$_{\pm 1.18}$ 
    & 26.85$_{\pm 1.92}$ \\

TFM 
    & 123.39
    & 10.12$_{\pm 0.13}$ 
    & 19.60$_{\pm 0.72}$ 
    & 22.41$_{\pm 0.82}$ \\

GFM 
    & 115.05
    & 9.75$_{\pm 0.23}$ 
    & 18.67$_{\pm 0.72}$ 
    & 21.55$_{\pm 0.54}$ \\

GiFlow 
    & 104.29
    & $\mathbf{9.54_{\pm 0.18}}$ 
    & $\mathbf{18.10_{\pm 0.78}}$ 
    & $\mathbf{21.27_{\pm 0.33}}$ \\
\bottomrule
\end{tabular}
}}
\end{table}
\begin{table}[t]
\caption{The performance of GiFlow with reduced components.}
\label{tab:ablation_components} 
\centering
{
\resizebox{1\columnwidth}{!}{
\begin{tabular}{lccc}
\toprule
 & MAE & RMSE & MAPE \\
\midrule
GiFlow & $9.54 _{\pm 0.18}$ & $18.10 _{\pm 0.78}$ & $21.27 _{\pm 0.33}$ \\
[0.2em]
\cdashline{1-4}\\[-0.8em]
w/o spatial attention & $10.05 _{\pm 0.21}$ & $19.97 _{\pm 0.68}$ & $21.46 _{\pm 0.52}$ \\
w/o temporal attention & $9.87 _{\pm 0.25}$ & $19.59 _{\pm 0.97}$ & $21.53 _{\pm 0.71}$ \\
w/o spatiotemporal attention & $10.22 _{\pm 0.22}$ & $20.02 _{\pm 0.85}$ & $21.88 _{\pm 0.75}$ \\
w/o spatiotemporal propagation & $10.40 _{\pm 0.81}$ & $19.92 _{\pm 1.06}$ & $23.22 _{\pm 2.44}$ \\
\bottomrule
\end{tabular}
}}
\end{table}

\textbf{Ablation on architectural components.}
We conduct ablation studies to examine the individual contributions of the vector field model components by evaluating four variants of GiFlow: removing spatial attention, removing temporal attention, removing both spatial and temporal attention, and removing spatiotemporal propagation.
In GiFlow, the outputs of the attention modules are concatenated and then projected to the hidden dimension through a linear layer. To remove an attention module, we simply exclude it from the concatenated representation and adjust the input dimension of the subsequent linear layer accordingly. To remove spatiotemporal propagation, we set the propagation layer $L_{MP}$ to zero.
The experiment is conducted on the Air-36 dataset under the point-missing setting with $\rho = 20\%$, and the results are reported in Table \ref{tab:ablation_components}. We observe that spatial attention, temporal attention, and spatiotemporal propagation each contribute positively to the overall performance. 
Notably, removing spatiotemporal propagation leads to the largest performance drop, highlighting the importance of information propagation in the spatiotemporal domain. Compared with the ablation study on the prior design, these results suggest that the prior design has a larger impact on performance than the architectural components.

\subsection{Sensitivity Analysis of Graph Quality}
In our experiments, the threshold for graph binarization is set to either 0.1 or 0.2.
To investigate how sensitive GiFlow is to the graph quality, we evaluate its performance using graphs constructed with different thresholds, ranging from 0.02 to 0.6. 
The experiment is conducted on the Air-36 dataset under the point-missing setting with $\rho = 20\%$, and we report the performance and the average degree of the resulting graphs in Table \ref{tab:graph_threshold} .
The results show that GiFlow performs consistently well when the threshold is between 0.05 and 0.4, with the best result at 0.1. However, under extreme threshold values, such as 0.02 or 0.6, the performance drops noticeably. These findings suggest that GiFlow is robust to moderate variations in graph quality, while its performance degrades significantly when the graph structure deviates substantially from a well-chosen one.
\begin{table}[t]
\caption{The performance of GiFlow with graphs generated by different thresholds.}
\label{tab:graph_threshold} 
\centering
{
\resizebox{1\columnwidth}{!}{
\begin{tabular}{ccccc}
\toprule
Threshold & Avg. degree & MAE & RMSE & MAPE \\
\midrule
0.02 & 22.22 & $10.33 _{\pm 0.57}$ & $20.07 _{\pm 0.82}$ & $22.34 _{\pm 1.48}$\\
0.05 & 19.27 & $9.97 _{\pm 0.20}$ & $19.86 _{\pm 0.83}$ & $22.19 _{\pm 0.61}$ \\
0.1 & 17.16 & $9.54 _{\pm 0.18}$ & $18.10 _{\pm 0.78}$ & $21.27 _{\pm 0.33}$ \\
0.2 & 14.11 & $9.74 _{\pm 0.17}$ & $18.44 _{\pm 0.27}$ & $21.41 _{\pm 0.63}$ \\
% 0.3 & 12.33 & 9.81 \pm 0.10 & 18.70 $\pm$ 0.72 & 21.78 $\pm$ 0.41 \\
0.4 & 10.55 & $9.94 _{\pm 0.18}$ & $19.01 _{\pm 0.86}$ & $21.84 _{\pm 0.60}$ \\
% 0.5 & 8.7 & 10.18 $\pm$ 0.85 & 19.33 $\pm$ 1.06 & 25.25 $\pm$ 2.08 \\
0.6 & 6.8 & $11.15 _{\pm 1.57}$ & $20.20 _{\pm 1.52}$ & $26.44 _{\pm 4.21}$\\
\bottomrule
\end{tabular}
}}
\end{table}

\subsection{Runtime Analysis}
\textbf{Inference time analysis.}
In this section, we evaluate the complexity of different generative models.
Specifically, we evaluate the inference time (in minutes) required to impute the test sets of the datasets for the considered diffusion models and GiFlow.
The experiments are conducted on an A100 NVIDIA GPU with 80GB of memory.
The results are reported in Table \ref{tab:runtime}.
From the results, it is evident that GiFlow significantly outperforms diffusion models in terms of inference speed.
\begin{table}[t]
\centering
\caption{The inference time on test set.}
\label{tab:runtime}
{
\resizebox{0.6\columnwidth}{!}{
\begin{tabular}{lcccc}
\toprule
Model & Air-36 & AQI & PeMS08 \\
\midrule
PriSTI & 9.30 & 43.12 & 7.46 \\
CoSTI & 0.37 & 8.41 & 3.63 \\
GiFlow & 0.28 & 2.47 & 0.99 \\
\bottomrule
\end{tabular}
}}
\end{table}

\textbf{Filtering factor optimization complexity analysis.}
During training, GiFlow requires the optimization of the adaptive filtering factors.
To evaluate the efficiency of this process, we optimize these filtering factors for 100 epochs with a batch size of 64. 
To evaluate the applicability of this optimization process to larger graphs, we generate synthetic graphs with 10,000 to 50,000 nodes using the same procedure described in Section \ref{sec:synthetic}. 
We report the total training time (in minutes) and peak GPU memory usage (in gigabytes) on both real-world datasets and large synthetic graphs in Table \ref{tab:filtering_factors}, where S-$i$ represents synthetic graphs with $i$ thousand nodes. 
The results validate that this optimization process is efficient and applicable to larger graphs.

    \begin{table}[t]
    \caption{The complexity of optimizing filtering factors.}
\label{tab:filtering_factors} 
    \centering
    {
    \resizebox{1\columnwidth}{!}{
    \begin{tabular}{lccccccc}
    \toprule
     & Air-36 & AQI & PeMS08 & S-10 & S-20 & S-30&S-50\\
    \midrule
    Time & 0.19 & 0.36 & 0.46 & 5.75 & 11.56 & 23.98 & 69.85 \\
    Memory & 0.02 & 0.04 & 0.03 & 1.01 & 3.49 & 7.47 & 19.88 \\
    \bottomrule
    \end{tabular}
    }}
    \end{table}
\textbf{Performance with fewer Euler steps.}
In our experiments, we use 20 Euler steps to do the imputation. To investigate how GiFlow performs with fewer generation steps, we evaluate its performance with different numbers of steps, ranging from 1 to 20.
We conduct experiments on the Air-36 dataset under the point-missing setting with $\rho = 20\%$. The performance and the inference time (in seconds) are shown in Table \ref{tab:eulerstep}.
The performance of GiFlow degrades slightly with decreased Euler steps. In particular, with 5 steps, GiFlow still achieves lower MAE and RMSE than the second-best baseline, while offering substantially faster inference. These results suggest that GiFlow provides a favorable efficiency--accuracy trade-off and remains effective even with fewer Euler steps.
\begin{table}[t]
\caption{Performance with different Euler steps.}
\label{tab:eulerstep}
\centering
{
\resizebox{1\columnwidth}{!}{
\begin{tabular}{ccccc}
\toprule
Euler steps & Inference time & MAE & RMSE & MAPE \\
\midrule
1 & 1.58 & $9.87 _{\pm 0.16}$ & $19.50 _{\pm 0.81}$ & $21.81 _{\pm 0.45}$\\
2 & 1.94 & $9.87 _{\pm 0.17}$ & $19.36 _{\pm 0.90}$ & $21.76 _{\pm 0.45}$ \\
5 & 4.60 & $9.81 _{\pm 0.15}$ & $18.95 _{\pm 0.81}$ & $21.63 _{\pm 0.45}$ \\
10 & 10.37 & $9.67 _{\pm 0.18}$ & $18.50 _{\pm 0.79}$ & $21.64 _{\pm 0.38}$ \\
20 & 17.15 & $9.54 _{\pm 0.18}$ & $18.10 _{\pm 0.78}$ & $21.27 _{\pm 0.33}$ \\
\bottomrule
\end{tabular}
}}
\end{table}

\section{Conclusion}
In this work, we consider the spatiotemporal imputation problem.
We developed a graph-informed flow matching method named GiFlow, which uses a graph-informed prior derived based on adaptive spatiotemporal filtering of observable signals.  
Compared with the problem-agnostic Gaussian prior, the proposed graph-informed prior better aligns with the target data distribution, and it provably reduces the transport cost from the source to the target distribution.
We also theoretically analyze the relationship between spatiotemporal filtering factors and the receptive field in the filtering process.
Experiments on both synthetic and real-world datasets with various missing patterns and missing rates demonstrate the effectiveness and robustness of the GiFlow model.

\section*{Acknowledgments}
This work was supported by the Swiss National Science Foundation (SNSF) under Grant No. 200021\_200461. The contributions of Aref Einizade and Jhony H. Giraldo to this project were supported by Hi! PARIS and the ANR/France 2030 program (Grant ANR-23-IACL-0005 and ANR-23-CMAS-0033).

\section*{Impact Statement}

This paper presents work whose goal is to advance the field of machine learning. There are many potential societal consequences of our work, none of which we feel must be specifically highlighted here.

\bibliography{iclr2026_conference}
\bibliographystyle{icml2026}

%%%%%%%%%%%%%%%%%%%%%%%%%%%%%%%%%%%%%%%%%%%%%%%%%%%%%%%%%%%%%%%%%%%%%%%%%%%%%%%
%%%%%%%%%%%%%%%%%%%%%%%%%%%%%%%%%%%%%%%%%%%%%%%%%%%%%%%%%%%%%%%%%%%%%%%%%%%%%%%
% APPENDIX
%%%%%%%%%%%%%%%%%%%%%%%%%%%%%%%%%%%%%%%%%%%%%%%%%%%%%%%%%%%%%%%%%%%%%%%%%%%%%%%
%%%%%%%%%%%%%%%%%%%%%%%%%%%%%%%%%%%%%%%%%%%%%%%%%%%%%%%%%%%%%%%%%%%%%%%%%%%%%%%
\newpage
\appendix
\onecolumn
\section{Related Work} \label{app:related-work}

%\textbf{Classical spatiotemporal imputation.}  
Spatiotemporal imputation addresses the problem of reconstructing missing values in data that combine temporal dynamics with spatial dependencies, with applications in domains such as air quality monitoring \citep{cao2018brits}, traffic forecasting \citep{li2018diffusion}, and weather prediction \citep{price2025gencast}. 
Early statistical approaches rely on distributional assumptions such as temporal smoothness or local similarity across series.
Examples include autoregressive models \citep{ansley1984estimation}, expectation-maximization \citep{nelwamondo2007missing}, and $k$-nearest neighbors \citep{beretta2016nearest}. 
While these methods are simple and theoretically well-understood, they struggle in real-world settings where spatial and temporal dependencies are highly nonlinear and heterogeneous.
Their limited capacity to capture complex interactions has motivated the development of more flexible machine learning approaches.

%\textbf{Deep learning and GNN-based models.}  
Neural methods have significantly advanced imputation by better capturing temporal and spatial dependencies. 
RNNs and their extensions exploit temporal correlations by recursively propagating hidden states, as in BRITS (bidirectional recurrent imputation for time series) \citep{cao2018brits}, while transformer-based designs such as SAITS (self-attention-based imputation for time series) \citep{du2023saits} introduce attention mechanisms to jointly optimize imputation and reconstruction.
However, these approaches typically operate on individual time series and ignore spatial relationships.
To address this limitation, GNN models have been proposed to incorporate spatial dependencies by propagating information over a graph topology.
Examples include GRIN (graph recurrent imputation network) \citep{cini2022filling}, SPIN (spatiotemporal point inference network) \citep{marisca2022learning}, and OPCR (one-step propagation and confidence-based refinement) \citep{deng2024learning}, which combine temporal modeling with graph-based message passing.
Despite their effectiveness, these iterative propagation schemes are prone to error accumulation and information bottlenecks, particularly under high missing rates \citep{cini2025graph}.

%\textbf{Generative models for imputation.}
Generative models provide an alternative paradigm by directly modeling conditional data distributions, thereby avoiding the accumulation of errors across propagation steps.
Diffusion-based approaches \citep{sohl2015deep,ho2020denoising,song2021scorebased} have demonstrated strong generative performance in multiple domains, from vision \citep{dhariwal2021diffusion,rombach2022high} to spatiotemporal data \citep{liu2023pristi,he2025filling}. 
Conditional diffusion frameworks, such as CSDI (conditional score-based diffusion models for imputation) \citep{tashiro2021csdi}, have been adapted to time-series imputation, while PriSTI (spatiotemporal imputation with enhanced prior modeling) \citep{liu2023pristi} and related models extend them to spatiotemporal settings.
However, these methods typically rely on problem-agnostic Gaussian priors and require computationally expensive iterative denoising.
In practice, accurate inference often demands multiple sampling runs followed by averaging, which limits both efficiency and robustness when applied to large-scale spatiotemporal data.

The iterative nature of diffusion models leads to high inference computational costs.
To improve the sampling efficiency of diffusion models, consistency models have been proposed, where the models are trained to map any point at any time step to the trajectory's starting point \citep{song2023consistency}.
The idea of the consistency model has also been applied to the problem of spatiotemporal imputation, which demonstrates significant improvement in inference computational costs \citep{solis2025costi}.
This line of approaches, however, still relies on the problem-agnostic Gaussian priors.
It has been shown that the idea of the consistency model can be extended to accommodate the flow matching framework as well \citep{liu2025see}.
But the effectiveness of such an extension has not been investigated yet in the spatiotemporal imputation task.

Flow matching \citep{albergo2023building, lipman2023flow, liu2023flow} generalizes diffusion by directly learning a continuous probability flow from a source to a target distribution, regressing vector fields along transport paths. 
Instead of relying on stochastic noise injection, flow matching directly learns a continuous probability flow that transports a source distribution to the target distribution. 
FM can accommodate arbitrary source distributions, although Gaussian priors are still often chosen for convenience. 
FM avoids stochastic noise injection, reduces training variance, and stabilizes optimization \citep{lipman2023flow, albergo2023building}. 
Moreover, the deterministic inference of FM enables efficient sampling without repeated averaging as required in diffusion models \citep{liu2023pristi}. 
These characteristics make FM particularly suitable for tasks where partial observations are available \citep{albergo2024stochastic,kollovieh2025flow,liu2025flowing}, since we can build problem-tailored priors based on it.
By aligning the prior distribution to the target distribution, the performance of FM can be further enhanced \citep{tong2024improving}.

\section{Proofs}\label{app:proofs}
\subsection{Proof of Proposition~\ref{prop:adaptive_receptive_field}}\label{sec:proof_prop}
Our results on the truncation error analysis of the spatiotemporal filtering can be viewed as an extension of the truncation error analysis for spatial filtering presented in \citep{behmanesh2023tide}.

The Taylor series for the spatiotemporal filtering defined in \eqref{eq:graph_filtering} is given by
\begin{equation}
\mathbf{X}_{\boldsymbol{\tau}}
= \left(\sum_{k=0}^{\infty}\frac{(-\tau_\eta)^k}{k!}\mathbf{L}_\eta^k\right) \mathbf{X}_1^M 
  \left(\sum_{m=0}^{\infty}\frac{(-\tau_\xi)^m}{m!}\mathbf{L}_\xi^m\right).
\end{equation}
Intuitively, for any nonzero $(\tau_\eta,\tau_\xi)$, this series propagates information from the whole graph, as no factor in front of the power of the Laplacian is zero.
The truncated version of $\mathbf{X}_{\boldsymbol{\tau}}$ is defined by
\begin{equation}
\mathbf{X}_{\boldsymbol{\tau}}^{K_\eta,K_\xi} =
\left(\sum_{k=0}^{K_\eta}\frac{(-\tau_\eta)^k}{k!}\mathbf{L}_\eta^k\right) \mathbf{X}_1^M
\left(\sum_{m=0}^{K_\xi}\frac{(-\tau_\xi)^m}{m!}\mathbf{L}_\xi^m\right).
\end{equation}
Therefore, we have
\begin{equation}
\begin{aligned}
&\quad\left\|\mathbf{X}_{\boldsymbol{\tau}} - \mathbf{X}_{\boldsymbol{\tau}}^{K_\eta,K_\xi}\right\| \\
&=   \Biggl\|\left(\sum_{k=0}^{\infty}\frac{(-\tau_\eta)^k}{k!}\mathbf{L}_\eta^k\right) \mathbf{X}_1^M
  \left(\sum_{m=0}^{\infty}\frac{(-\tau_\xi)^m}{m!}\mathbf{L}_\xi^m\right)-
  \left(\sum_{k=0}^{K_\eta}\frac{(-\tau_\eta)^k}{k!}\mathbf{L}_\eta^k\right) \mathbf{X}_1^M 
\left(\sum_{m=0}^{K_\xi}\frac{(-\tau_\xi)^m}{m!}\mathbf{L}_\xi^m\right)\Biggr\|\\
&=\Biggl\|\left(\left(\sum_{k=0}^{K_\eta}\frac{(-\tau_\eta)^k}{k!}\mathbf{L}_\eta^k\right) +\left(\sum_{k=K_\eta+1}^{\infty}\frac{(-\tau_\eta)^k}{k!}\mathbf{L}_\eta^k\right)\right)\mathbf{X}_1^M
\left(\sum_{m=0}^{\infty}\frac{(-\tau_\xi)^m}{m!}\mathbf{L}_\xi^m\right)-\\
&\quad\left(\sum_{k=0}^{K_\eta}\frac{(-\tau_\eta)^k}{k!}\mathbf{L}_\eta^k\right) \mathbf{X}_1^M
\left(\sum_{m=0}^{K_\xi}\frac{(-\tau_\xi)^m}{m!}\mathbf{L}_\xi^m\right)\Biggr\|\\
&=\Biggl\|\left(\sum_{k=K_\eta+1}^{\infty}\frac{(-\tau_\eta)^k}{k!}\mathbf{L}_\eta^k\right)\mathbf{X}_1^M
\left(\sum_{m=0}^{\infty}\frac{(-\tau_\xi)^m}{m!}\mathbf{L}_\xi^m\right)+\\
&\quad\quad\left(\sum_{k=0}^{K_\eta}\frac{(-\tau_\eta)^k}{k!}\mathbf{L}_\eta^k\right) \mathbf{X}_1^M
\left(\sum_{m=K_\xi+1}^{\infty}\frac{(-\tau_\xi)^m}{m!}\mathbf{L}_\xi^m\right)\Biggr\|\\
&\leq \Biggl\|\left(\sum_{k=K_\eta+1}^{\infty}\frac{(-\tau_\eta)^k}{k!}\mathbf{L}_\eta^k\right)\mathbf{X}_1^M 
\left(\sum_{m=0}^{\infty}\frac{(-\tau_\xi)^m}{m!}\mathbf{L}_\xi^m\right)\Biggr\|+\\
&\quad\quad\Biggl\|
\left(\sum_{k=0}^{K_\eta}\frac{(-\tau_\eta)^k}{k!}\mathbf{L}_\eta^k\right) \mathbf{X}_1^M
\left(\sum_{m=K_\xi+1}^{\infty}\frac{(-\tau_\xi)^m}{m!}\mathbf{L}_\xi^m\right)\Biggr\|\\
&\leq \left(\left(\sum_{k=K_\eta+1}^{\infty}\frac{|\tau_\eta|^k}{k!} C_s^k \right)
\left(\sum_{m=0}^{\infty}\frac{|\tau_\xi|^m}{m!} C_t^m \right)+
\left(\sum_{k=0}^{\infty}\frac{|\tau_\eta|^k}{k!} C_s^k \right)
\left(\sum_{m=K_\xi+1}^{\infty}\frac{|\tau_\xi|^m}{m!} C_t^m \right)\right) \|\mathbf{X}_1^M\|,
\end{aligned}
\end{equation}
which completes the proof.

\subsection{Proof of Theorem~\ref{thm:transport-cost}}\label{sec:proof_transport_cost}
Let $\mathbf{X}_{0}\sim p_{0}$ and $\mathbf{X}_{1}\sim q_{1}$ be the source and target distributions of a flow matching model, while $\mathbf{X}_1^M=\mathbf{X}_1\circ\mathbf{M}$ being the observable data, then the transport cost $\mathcal{C}_{\mathrm{FM}}(p_0 \to q_1)$ is defined as follows:
\begin{equation}
    \mathcal{C}_{\mathrm{FM}}(p_0 \to q_1)=\mathbb{E}_{\mathbf{X}_{1}\sim p_{1}(\mathbf{X}_{0})}\left[\left\Vert \mathbf{X}_{t=0}(\mathbf{X}_{1}^M)-\mathbf{X}_{1}\right\Vert ^{2}\right],
\end{equation}
where $\mathbf{X}_{t=0}(\mathbf{X}_{1}^M)$ represents the source sample generated from the observable data $\mathbf{X}_{1}^M$.
Specifically, for FM with a Gaussian prior, we have $\mathbf{X}_{t=0}^{Gauss}(\mathbf{X}_{1}^M)=\mathbf{X}_{1}^M+\boldsymbol{\Sigma}\circ\left(\mathbf{1}-\mathbf{M}\right)$ with $\boldsymbol{\Sigma}\sim\mathcal{N}\left(\mathbf{0},\sigma^2\mathbf{I}\right)$ sampled from an isotropic Gaussian distribution.
Therefore, we have
\begin{equation}
\begin{aligned}
    \mathcal{C}_{\mathrm{FM}}(p_0^{Gauss} \to q_1)&=\mathbb{E}_{\mathbf{X}_{1}\sim p_{1}(\mathbf{X}_{0})}\left[\left\Vert \mathbf{X}_{1}^M+\boldsymbol{\Sigma}\circ\left(\mathbf{1}-\mathbf{M}\right)-\mathbf{X}_{1}\right\Vert ^{2}\right]\\
    &=\mathbb{E}_{\mathbf{X}_{1}\sim p_{1}(\mathbf{X}_{0})}\left[\left\Vert \mathbf{X}_{1}\circ\mathbf{M}+\boldsymbol{\Sigma}\circ\left(\mathbf{1}-\mathbf{M}\right)-\mathbf{X}_{1}\right\Vert ^{2}\right]\\
    &=\mathbb{E}_{\mathbf{X}_{1}\sim p_{1}(\mathbf{X}_{0})}\left[\left\Vert \left(\mathbf{X}_{1}-\boldsymbol{\Sigma}\right)\circ\left(\mathbf{1}-\mathbf{M}\right)\right\Vert ^{2}\right]\\
    &=\mathbb{E}_{\mathbf{X}_{1}\sim p_{1}(\mathbf{X}_{0})}\left[\left\Vert \mathbf{X}_{1}\circ\left(\mathbf{1}-\mathbf{M}\right)\right\Vert ^{2}\right]+\mathbb{E}_{\mathbf{X}_{1}\sim p_{1}(\mathbf{X}_{0})}\left[\left\Vert \boldsymbol{\Sigma}\circ\left(\mathbf{1}-\mathbf{M}\right)\right\Vert ^{2}\right]\\
    &=\mathbb{E}_{\mathbf{X}_{1}\sim p_{1}(\mathbf{X}_{0})}\left[\left\Vert \mathbf{X}_{1}\circ\left(\mathbf{1}-\mathbf{M}\right)\right\Vert ^{2}\right]+\sigma^2\mathrm{tr}\left(\mathbf{1}-\mathbf{M}\right).
\end{aligned}
\label{eq:gauss_transport_cost}
\end{equation}
For GiFlow, the graph-informed prior is constructed based on the spatiotemporal filtering defined in \eqref{eq:graph_filtering}, leading to $\mathbf{X}_{t=0}^G(\mathbf{X}_{1}^M)=e^{-\tau_\eta \mathbf{L}_\eta} \, \mathbf{X}_{1}^M \, e^{-\tau_\xi \mathbf{L}_\xi}$.
Since $\tau_\eta$ and $\tau_\xi$ are obtained by optimizing \eqref{eq:tau-optimization} with $\alpha_\tau=0$, we have
\begin{equation}
    \left\Vert \mathbf{X}_{1}-e^{-\tau_\eta\mathbf{L}_\eta}\mathbf{X}_1^Me^{-\tau_\xi\mathbf{L}_\xi}\right\Vert^2\leq\left\Vert \mathbf{X}_{1}-\mathbf{X}_1^M\right\Vert^2
\label{eq:optimal_objective}
\end{equation}
since $\tau_\eta$ and $\tau_\xi$ represent the optimal solution of \eqref{eq:tau-optimization}.
Note that since Problem \eqref{eq:tau-optimization} is nonconvex, we can only obtain stationary solutions in practice.
In such cases, we can optimize \eqref{eq:tau-optimization} using gradient-based methods with initialization $\tau_\eta=0$ and $\tau_\xi=0$, then Eq. \eqref{eq:optimal_objective} still holds.
Therefore, we have
\begin{equation}
\begin{aligned}
    \mathcal{C}_{\mathrm{FM}}(p_0^{G} \to q_1)&=\mathbb{E}_{\mathbf{X}_{1}\sim p_{1}(\mathbf{X}_{0})}\left[\left\Vert \mathbf{X}_{1}-e^{-\tau_\eta\mathbf{L}_\eta}\mathbf{X}_1^Me^{-\tau_\xi\mathbf{L}_\xi}\right\Vert ^{2}\right]\\
    &\leq\mathbb{E}_{\mathbf{X}_{1}\sim p_{1}(\mathbf{X}_{0})}\left[\left\Vert \mathbf{X}_{1}-\mathbf{X}_1^M\right\Vert ^{2}\right]\\
    &=\mathbb{E}_{\mathbf{X}_{1}\sim p_{1}(\mathbf{X}_{0})}\left[\left\Vert \mathbf{X}_{1}-\mathbf{X}_1\circ\mathbf{M}\right\Vert ^{2}\right]\\
    &\leq\mathbb{E}_{\mathbf{X}_{1}\sim p_{1}(\mathbf{X}_{0})}\left[\left\Vert \mathbf{X}_{1}\circ\left(\mathbf{1}-\mathbf{M}\right)\right\Vert ^{2}\right]+\sigma^2\mathrm{tr}\left(\mathbf{1}-\mathbf{M}\right).
\end{aligned}
\label{eq:gifm_transport_cost}
\end{equation}
Combining Eq. \eqref{eq:gauss_transport_cost} and Eq. \eqref{eq:gifm_transport_cost} completes the proof.

\section{Experiments}
\subsection{Introduction of Datasets} \label{app:datasets}
We conduct experiments on three real-world datasets: two air quality datasets, Air-36 and AQI, and a traffic dataset, PeMS08.
Air-36 and AQI collect hourly-sampled PM2.5 pollutant data.
Specifically, Air-36 is collected from 36 monitoring stations in Beijing, while AQI is collected from 437 monitoring stations spread across 43 Chinese cities.
Both air quality datasets have 8760 timesteps, covering one year from 2014/05/01 to 2015/04/30 \citep{zheng2015forecasting}.
PeMS08 is a traffic dataset about highway traffic flow in California, which is collected by the Caltrans Performance Measurement System (PeMS) \citep{chen2001freeway}.
Specifically, it is collected from 170 monitoring sensors covering two months from 2016/07/01 to 2016/08/31.
It originally collects data every 30 seconds, and the collected data is then aggregated with a 5-minute interval.
The statistics of the datasets are summarized in Table \ref{tab:dataset_statistics}.

\begin{table}[h]
\centering
\caption{Statistics of the datasets.}
\label{tab:dataset_statistics}
\begin{tabular}{lcccc}
\toprule
Dataset & \# Nodes & \# Timesteps & Sampling intervals & Collected date \\
\midrule
Air-36 & 36 & 8760 & 1 hour & 2014/05/01 -- 2015/04/30 \\
AQI & 437 & 8760 & 1 hour & 2014/05/01 -- 2015/04/30 \\
PeMS08 & 170 & 17856 & 5 minute & 2016/07/01 -- 2016/08/31 \\
\bottomrule
\end{tabular}
\end{table}

\subsection{Introduction of Baselines} \label{app:baselines}
To evaluate the performance of our proposed method, we compare it with various baselines:

\begin{itemize}
    \item \textbf{Mean-S}: imputes the missing values using the spatial average, i.e., the mean values of all nodes at a given timestep.
    \item \textbf{Mean-T}: imputes the missing values using the temporal average, i.e., the mean values of all timesteps for a given node.
    \item \textbf{Linear}: imputes the missing values using temporal linear interpolation for each node independently.
    \item \textbf{KNN}: imputes the missing values using the average signal of neighboring nodes.
    \item \textbf{FP} \citep{rossi2022on}: performs feature propagation to impute the missing values.
    \item \textbf{BRITS} \citep{cao2018brits}: a bidirectional RNN-based model.
    \item \textbf{SAITS} \citep{du2023saits}: a transformer-based model.
    \item \textbf{SPIN} \citep{marisca2022learning}: an efficient version of spatio-temporal attention-based method.
    \item \textbf{GRIN} \citep{cini2022filling}: a GNN model with bidirectional gated recurrent unit.
    \item \textbf{OPCR} \citep{deng2024learning}: a GNN model that contains attention-based one-step propagation and confidence-based refinement.
    \item \textbf{PriSTI} \citep{liu2023pristi}: a conditional diffusion model.
    \item \textbf{CoSTI} \citep{solis2025costi}: a consistency model. 
\end{itemize}

\subsection{Implementation Details}\label{app:implementation_details}
For all the experimental results, we give the average performance and standard deviation with 5 independent trials.
For all the datasets, we select windows of length 24.
For each dataset, we randomly select 70\%/10\%/20\% of the data for training, validation, and testing. 
The Adam optimizer is used in all experiments for model training \citep{KingBa15}.
We fix the maximum number of epochs to 300, and we use early stopping on the validation set with a patience of 10 epochs.
To stabilize the training process, we employed an exponential moving average (EMA) of the model parameters with a decay rate of 0.9999.
To solve the ODE, we utilized an Euler solver with 20 steps.
The models' hyperparameters are tuned based on the results of the validation set.
The search space of hyperparameters are as follows:
1) learning rate: \{0.005, 0.001, 0.0005\}; 2) weight decay: \{0, 5e-4, 5e-5, 5e-3\}; 3) dropout rate: \{0, 0.1, 0.2, 0.3\}; 4) GNN layers: \{2, 4, 6, 8\}; 5) embedding dimension: \{32, 64, 128\}; 6) weight parameter $\alpha_\tau$: \{0.1, 0.01, 0.001, 0.0001\}.

\subsection{Performance with different missing rate}\label{app:missing_rate}
\begin{figure}[b]
    \centering
    \includegraphics[width=0.55\textwidth]{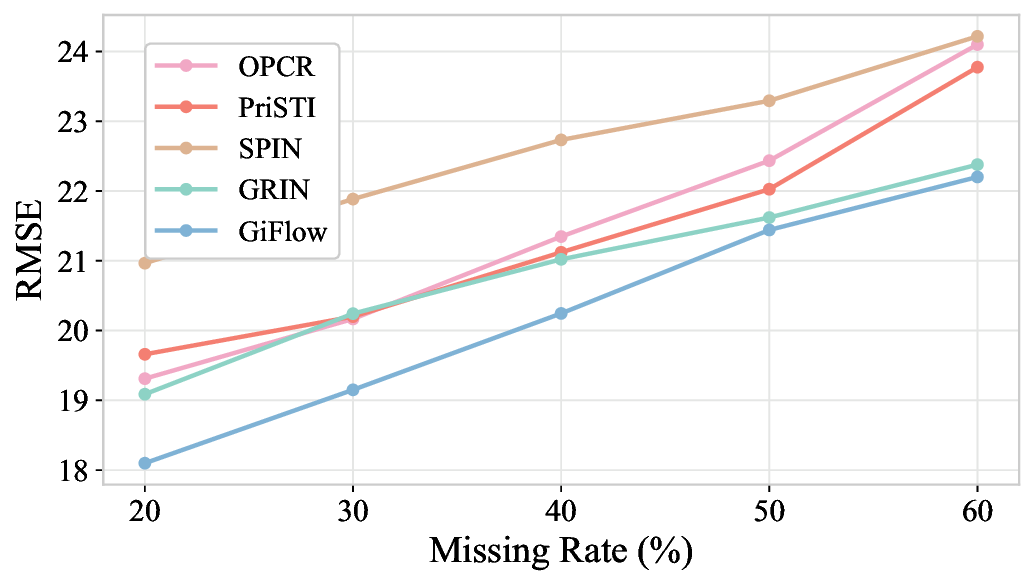}
    \caption{Performance on RMSE with different Missing Rate.}
    \label{fig:missing_rate_rmse}
\end{figure}
To evaluate the robustness of the model under different missing rates, we conduct experiments using the point missing strategy on the Air-36 dataset, with $\rho$ ranging from 20\% to 60\%.
The results on MAE and MAPE are showcased in Figure \ref{fig:missing-rate}.
In the following, we present the results on RMSE in Figure \ref{fig:missing_rate_rmse}.
The results on RMSE exhibit similar patterns as in the MAE and MAPE results, where the imputation performance steadily degrades with increasing rates, and GiFlow consistently outperforms the other baselines across different missing rates.

\subsection{Filtering factor values with increasing missing rate}\label{app:tau_value}
\begin{figure}[h]
    \centering
    \includegraphics[width=0.55\textwidth]{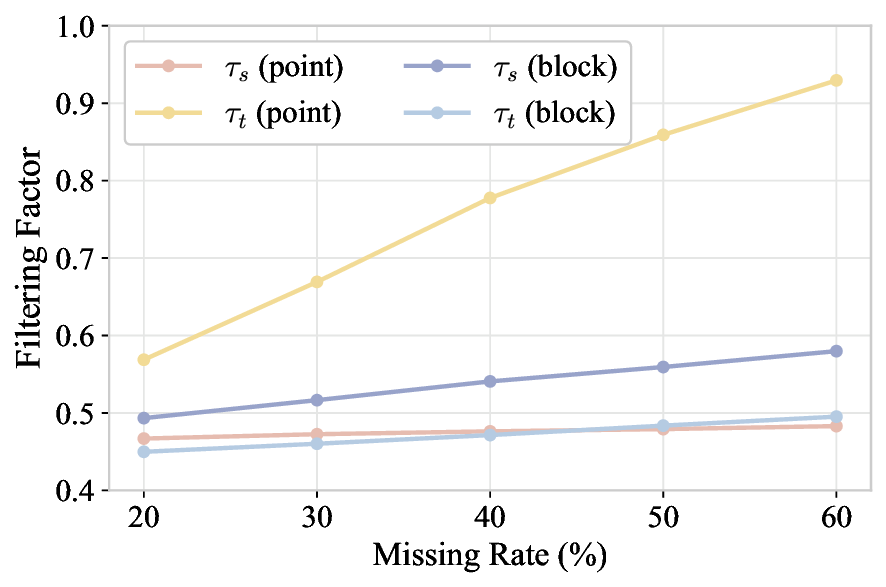}
    \caption{Filtering factor values with different missing rates on the AQI dataset.}
    \label{fig:tau_AQI}
\end{figure}

To validate the theoretical results presented in Proposition \ref{prop:adaptive_receptive_field}, we evaluate how the filtering factors $\tau_\eta$ and $\tau_\xi$ change with increasing missing rates.
The results on the Air-36 dataset are showcased in Figure \ref{fig:tau-missing}.
In the following, we present the results on the AQI dataset in Figure \ref{fig:tau_AQI}.
The patterns of $\tau_\eta$ and $\tau_\xi$ with increasing missing rate on the AQI dataset coincide with the patterns on the Air-36 dataset.
Specifically, the filtering factors $\tau_\eta$ and $\tau_\xi$ increase as the missing rate increases, which corroborates the results we obtained from Proposition \ref{prop:adaptive_receptive_field}.
Moreover, we observe that the increase of $\tau_\eta$ is more significant than the increase of $\tau_\xi$, indicating that the model relies more on spatial filtering than temporal filtering. 

\subsection{Applicability to Other Datasets}
In this section, we evaluate GiFlow on a traffic dataset about highway traffic flow in California to validate its applicability to other datasets.
Specifically, we conduct experiments using both the point missing strategy and block missing strategy with $\rho=20\%$ on the PeMS08 dataset.
The results are reported in Table \ref{tab:traffic}.
It can be observed that KNN and FP perform quite bad, indicating that relying only on the spatial dependencies cannot characterize the system dynamics well.
The spatiotemporal methods still achieve good results, highlighting the importance of considering both spatial and temporal dependencies.
The proposed GiFlow achieves the best results on all metrics, validating the applicability of GiFlow to other datasets under different missing patterns.

\begin{table*}[h]
\centering
\caption{Imputation performance on PeMS08 under point and block missing strategies.}
\label{tab:traffic}

\resizebox{0.85\textwidth}{!}{
\begin{tabular}{lcccccc}
\toprule
\multirow{2}{*}{Model} 
& \multicolumn{3}{c}{Point Missing} 
& \multicolumn{3}{c}{Block Missing} \\
\cmidrule(lr){2-4} \cmidrule(lr){5-7}
 & MAE & RMSE & MAPE & MAE & RMSE & MAPE \\
\midrule

Mean-S  & 86.64$_{\pm 0.11}$ & 113.59$_{\pm 0.12}$ & 141.40$_{\pm 1.67}$
        & 34.69$_{\pm 0.25}$ & 55.86$_{\pm 0.67}$ & 32.39$_{\pm 2.89}$ \\

Mean-T  & 21.27$_{\pm 0.05}$ & 39.74$_{\pm 0.07}$ & 13.96$_{\pm 0.06}$
        & 86.36$_{\pm 0.47}$ & 113.37$_{\pm 0.57}$ & 141.93$_{\pm 3.10}$ \\

Linear  & 14.71$_{\pm 0.04}$ & 24.04$_{\pm 0.08}$ & 9.94$_{\pm 0.05}$
        & 32.17$_{\pm 0.32}$ & 56.26$_{\pm 0.47}$ & 34.51$_{\pm 3.03}$ \\

KNN     & 117.65$_{\pm 0.19}$ & 152.50$_{\pm 0.22}$ & 195.18$_{\pm 2.10}$
        & 117.40$_{\pm 1.13}$ & 152.28$_{\pm 1.03}$ & 195.49$_{\pm 6.17}$ \\

FP      & 118.93$_{\pm 0.08}$ & 149.14$_{\pm 0.30}$ & 206.13$_{\pm 2.96}$
        & 118.97$_{\pm 0.44}$ & 149.08$_{\pm 0.34}$ & 206.26$_{\pm 5.25}$ \\

BRITS   & 17.66$_{\pm 0.03}$ & 29.13$_{\pm 0.08}$ & 12.07$_{\pm 0.10}$
        & 24.29$_{\pm 0.25}$ & 38.36$_{\pm 0.79}$ & 20.45$_{\pm 1.62}$ \\

SAITS   & 18.15$_{\pm 0.13}$ & 28.28$_{\pm 0.17}$ & 12.48$_{\pm 0.27}$
        & 32.31$_{\pm 1.82}$ & 47.23$_{\pm 2.22}$ & 24.55$_{\pm 2.27}$ \\

SPIN    & 14.82$_{\pm 0.06}$ & 24.26$_{\pm 0.43}$ & 9.34$_{\pm 0.06}$
        & 19.63$_{\pm 0.16}$ & \underline{30.43$_{\pm 0.33}$} & 14.94$_{\pm 1.85}$ \\

GRIN    & 13.72$_{\pm 0.12}$ & 21.49$_{\pm 0.29}$ & 8.82$_{\pm 0.27}$
        & 21.65$_{\pm 0.41}$ & 36.28$_{\pm 0.49}$ & 16.42$_{\pm 1.37}$ \\

OPCR    & \underline{12.77$_{\pm 0.20}$} & \underline{19.88$_{\pm 0.32}$} & \underline{8.67$_{\pm 0.52}$}
        & 22.96$_{\pm 1.06}$ & 39.89$_{\pm 1.13}$ & 21.70$_{\pm 2.98}$ \\

PriSTI  & 13.02$_{\pm 0.37}$ & 20.08$_{\pm 0.45}$ & 8.79$_{\pm 0.86}$
        & \underline{18.94$_{\pm 1.01}$} & 30.77$_{\pm 1.32}$ & \underline{14.48$_{\pm 0.54}$} \\
\addlinespace[0.2em]
\cdashline{1-7}
\addlinespace[0.2em]
GiFlow  & $\mathbf{12.66_{\pm 0.19}}$ & $\mathbf{19.83_{\pm 0.05}}$ & $\mathbf{8.43_{\pm 0.18}}$
        & $\mathbf{18.70_{\pm 0.37}}$ & $\mathbf{29.97_{\pm 0.37}}$ & $\mathbf{14.02_{\pm 1.67}}$ \\
\bottomrule
\end{tabular}
}
\end{table*}

%%%%%%%%%%%%%%%%%%%%%%%%%%%%%%%%%%%%%%%%%%%%%%%%%%%%%%%%%%%%%%%%%%%%%%%%%%%%%%%
%%%%%%%%%%%%%%%%%%%%%%%%%%%%%%%%%%%%%%%%%%%%%%%%%%%%%%%%%%%%%%%%%%%%%%%%%%%%%%%
\newpage
\end{document}